\documentclass{article}

\PassOptionsToPackage{square, compress, numbers}{natbib}

\usepackage[preprint]{neurips_2023}
\usepackage{multicol}
\usepackage{multirow}
\usepackage{graphicx}
\usepackage{floatrow}
\usepackage{float}
\usepackage{lipsum}  
\floatstyle{plaintop}
\restylefloat{table}
\usepackage{wrapfig}
\usepackage{array}
\usepackage[tableposition=top]{caption}
\newcolumntype{M}[1]{>{\centering\arraybackslash}m{#1}}

\usepackage{hyperref}       
\usepackage{booktabs}       
\usepackage{microtype}      
\usepackage{makecell}

\title{One-2-3-45: Any Single Image to 3D Mesh in 45 Seconds without Per-Shape Optimization}

\author{%
\textbf{Minghua Liu$^{1}$}\thanks{Equal Contribution} \and
\textbf{Chao Xu$^{2*}$} \and
\textbf{Haian Jin$^{3,4*}$} \and
\textbf{Linghao Chen$^{1,4*}$} \and
\textbf{Mukund Varma T$^{5}$} \and
\textbf{Zexiang Xu$^{6}$} \and
\textbf{Hao Su$^{1}$} 
\\
\\
$^{1}$ UC San Diego \ 
$^{2}$ UCLA \  
${^3}$ Cornell University \ 
${^4}$ Zhejiang University \ 
${^5}$ IIT Madras \ 
${^6}$ Adobe 
\\
\\
Project Website: \url{http://one-2-3-45.com}
}

\begin{document}

\maketitle
\vspace{-2em}

\begin{abstract}
\vspace{-0.5em}
Single image 3D reconstruction is an important but challenging task that requires extensive knowledge of our natural world. Many existing methods solve this problem by optimizing a neural radiance field under the guidance of 2D diffusion models but suffer from lengthy optimization time, 3D inconsistency results, and poor geometry. In this work, we propose a novel method that takes a single image of any object as input and generates a full 360-degree 3D textured mesh in a single feed-forward pass. Given a single image, we first use a view-conditioned 2D diffusion model, Zero123, to generate multi-view images for the input view, and then aim to lift them up to 3D space. Since traditional reconstruction methods struggle with inconsistent multi-view predictions, we build our 3D reconstruction module upon an SDF-based generalizable neural surface reconstruction method and propose several critical training strategies to enable the reconstruction of 360-degree meshes. Without costly optimizations, our method reconstructs 3D shapes in significantly less time than existing methods. Moreover, our method favors better geometry, generates more 3D consistent results, and adheres more closely to the input image. We evaluate our approach on both synthetic data and in-the-wild images and demonstrate its superiority in terms of both mesh quality and runtime. In addition, our approach can seamlessly support the text-to-3D task by integrating with off-the-shelf text-to-image diffusion models.
\vspace{-1em}
\end{abstract}

\vspace{-0.7em}
\section{Introduction}
\vspace{-0.7em}

\begin{figure}[t]
    \centering
    \includegraphics[width=\linewidth]{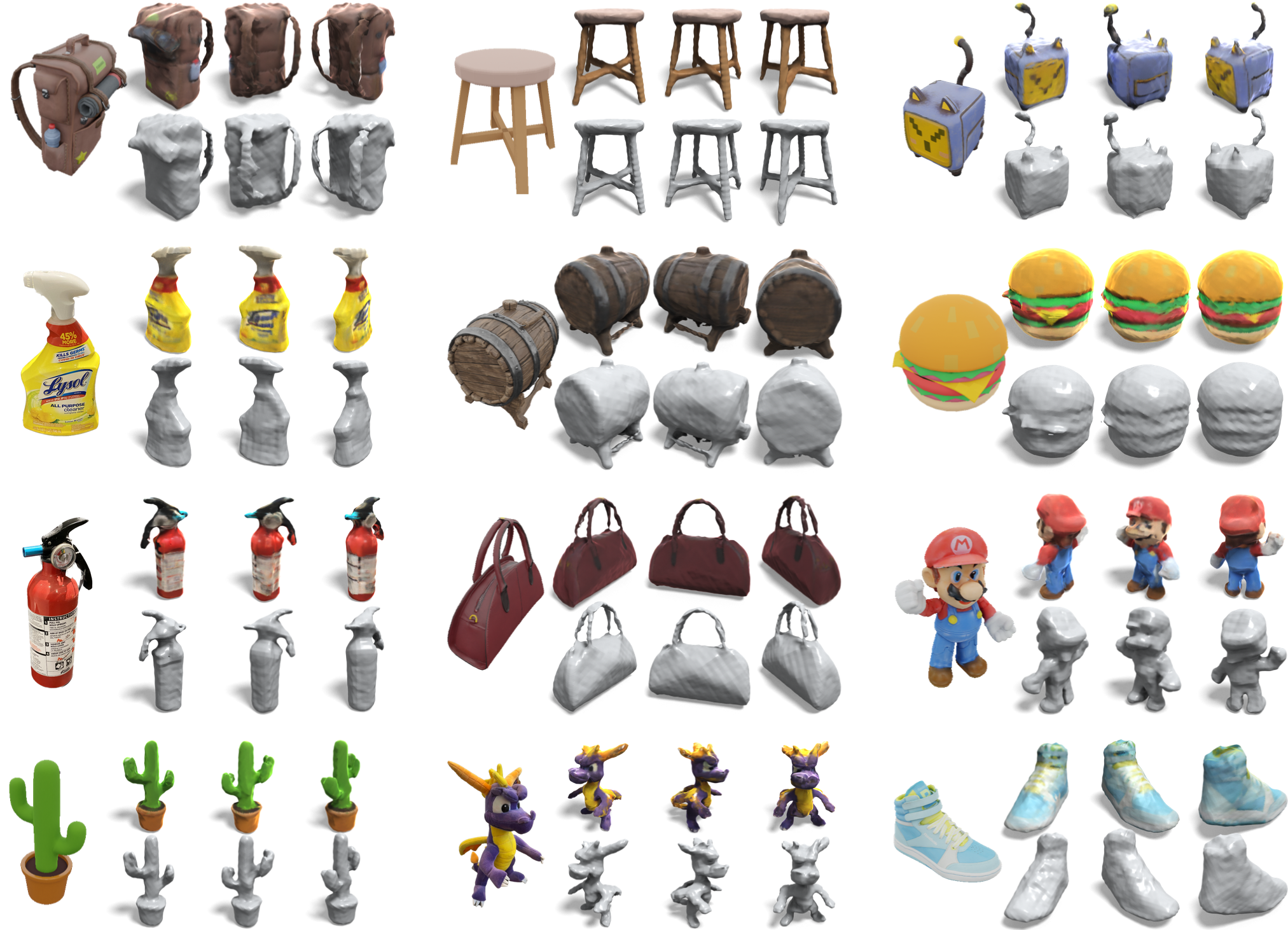}
    \vspace{-2em}
    \caption{One-2-3-45 reconstructs a full $360^{\circ}$  mesh of any object in 45 seconds given a single image of it. In each example, we showcase the input image in the left column, alongside the generated textured and textureless meshes from three different views. \vspace{-1.5em}}
    \label{fig:teaser}
\end{figure}

Single image 3D reconstruction, the task of reconstructing a 3D model of an object from a single 2D image, is a long-standing problem in the computer vision community and is crucial for a wide range of applications, such as robotic object manipulation and navigation, 3D content creation, as well as AR/VR~\cite{mees2019self,chiu2009automatic,yang2021robotic}. The problem is challenging as it requires not only the reconstruction of visible parts but also the hallucination of invisible regions. Consequently, this problem is often ill-posed and corresponds to multiple plausible solutions because of insufficient evidence from a single image. On the other hand, humans can adeptly infer unseen 3D content based on our extensive knowledge of the 3D world.
To endow intelligent agents with this ability, many existing methods~\cite{kanazawa2018learning,girdhar2016learning,huang2022planes,choy20163d,xie2020pix2vox++,yagubbayli2021legoformer,fan2017point,wen2019pixel2mesh++} exploit class-specific priors by training 3D generative networks on 3D shape datasets~\cite{chang2015shapenet}. However, these methods often fail to generalize to unseen categories, and their reconstruction quality is constrained by the limited size of public 3D datasets.

In this work, we pursue a generic solution to turn an image of any object, regardless of its category, into a high-quality 3D textured mesh. To achieve this,  we propose a novel approach that can effectively utilize the strong priors learned by 2D diffusion models for 3D reconstruction. Compared to 3D data, 2D images are more readily available and scalable. Recent 2D generative models (\eg, DALL-E~\cite{ramesh2021zero,ramesh2022hierarchical}, Imagen~\cite{saharia2022photorealistic}, and Stable Diffusion~\cite{rombach2022high}) and visual-language models (\eg, CLIP~\cite{radford2021learning}) have made significant strides by pre-training on Internet-scale image datasets. Since they learn a wide range of visual concepts and possess strong priors about our 3D world, it is natural to marry 3D tasks with them. Consequently, an emerging body of research~\cite{jain2022zero,hong2022avatarclip,michel2022text2mesh,poole2022dreamfusion,lin2022magic3d}, as exemplified by DreamField~\cite{jain2022zero}, DreamFusion~\cite{poole2022dreamfusion}, and Magic3D~\cite{lin2022magic3d}, employs 2D diffusion models or vision language models to assist 3D generative tasks. The common paradigm of them is to perform per-shape optimization with differentiable rendering and the guidance of the CLIP model or 2D diffusion models. While many other 3D representations have been explored, neural fields are the most commonly used representation during optimization.

Although these optimization-based methods have achieved impressive results on both text-to-3D~\cite{poole2022dreamfusion,jain2022zero,lin2022magic3d} and image-to-3D tasks~\cite{melas2023realfusion,seo2023let}, they face some common dilemmas: (a) \textbf{time-consuming}. Per-shape optimization typically involves tens of thousands of iterations of full-image volume rendering and prior model inferences, resulting in typically tens of minutes per shape. (b) \textbf{memory intensive}. Since the full image is required for the 2D prior model, the volume rendering can be memory-intensive when the image resolution goes up. (c) \textbf{3D inconsistent}. Since the 2D prior model only sees a single view at each iteration and tries to make every view look like the input, they often generate 3D inconsistent shapes (\eg, with two faces, or the Janus problem~\cite{melas2023realfusion,poole2022dreamfusion}). (d) \textbf{poor geometry}. Many methods utilize the density field as the representation in volume rendering. It is common that they produce good RGB renderings but extracting high-quality mesh tends to be difficult.

In this paper, instead of following the common optimization-based paradigm, we propose a novel approach to utilize 2D prior models for 3D modeling. At the heart of our approach is the combination of a 2D diffusion model with a cost-volume-based 3D reconstruction technique, enabling the reconstruction of a high-quality 360$^\circ$ textured mesh from a single image in a feed-forward pass without per-scene optimization. Specifically, we leverage a recent 2D diffusion model, Zero123~\cite{liu2023zero}, which is fine-tuned on Stable Diffusion~\cite{rombach2022high} to predict novel views of the input image given the camera transformation. We utilize it to generate multi-view predictions of the input single image so that we can leverage multi-view 3D reconstruction techniques to obtain a 3D mesh. There are two challenges associated with reconstruction from synthesized multi-view predictions: (a) the inherent lack of perfect consistency within the multi-view predictions, which can lead to severe failures in optimization-based methods such as NeRF methods \cite{mildenhall2021nerf,chen2022tensorf}. (b) the camera pose of the input image is required but unknown. To tackle them, we build our reconstruction module upon a cost volume-based neural surface reconstruction approach, SparseNeuS~\cite{long2022sparseneus}, which is a variant of MVSNeRF~\cite{chen2021mvsnerf}. Additionally, we introduce a series of essential training strategies that enable the reconstruction of 360-degree meshes from inherently inconsistent multi-view predictions. We also propose an elevation estimation module that estimates the elevation of the input shape in Zero123's canonical coordinate system, which is used to compute the camera poses required by the reconstruction module.

By integrating the three modules of multi-view synthesis, elevation estimation, and 3D reconstruction, our method can reconstruct 3D meshes of any object from a single image in a feed-forward manner. Without costly optimizations, our method reconstructs 3D shapes in significantly less time, \eg, in just 45 seconds. Our method favors better geometry due to the use of SDF representations, and generates more consistent 3D meshes, thanks to the camera-conditioned multi-view predictions. Moreover, our reconstruction adheres more closely to the input image compared to existing methods. See Figure~\ref{fig:teaser} for some of our example results. We evaluate our method on both synthetic data and real images and demonstrate that our method outperforms existing methods in terms of both quality and efficiency.

\vspace{-0.8em}
\section{Related Work}

\vspace{-0.8em}
\subsection{3D Generation Guided by 2D Prior Models}
\vspace{-0.6em}

Recently, 2D generative models (\eg, DALL-E~\cite{ramesh2021zero,ramesh2022hierarchical}, Imagen~\cite{saharia2022photorealistic}, and Stable Diffusion~\cite{rombach2022high}) and vision-language models (\eg, CLIP~\cite{radford2021learning}) have learned a wide range of visual concepts by pre-training on Internet-scale image datasets. They possess powerful priors about our 3D world and have inspired a growing body of research to employ 2D prior models for assisting 3D generative tasks. Exemplified by DreamField~\cite{jain2022zero}, DreamFusion~\cite{poole2022dreamfusion}, and Magic3D~\cite{lin2022magic3d}, a line of works follows the paradigm of per-shape optimization. They typically optimize a 3D representation (\ie, NeRF, mesh, SMPL human model) and utilize differentiable rendering to generate 2D images from various views. The images are then fed to the CLIP model~\cite{hong2022avatarclip,jain2022zero,michel2022text2mesh,lee2022understanding,canfes2023text,khalid2022text,aneja2022clipface,jetchev2021clipmatrix,xu2022dream3d,liu2022iss} or 2D diffusion model~\cite{poole2022dreamfusion,lin2022magic3d,seo2023let,melas2023realfusion,deng2022nerdi,wang2022score,xu2022neurallift,metzer2022latent,zhou2023sparsefusion,raj2023dreambooth3d} for calculating the loss functions, which are used to guide the 3D shape optimization. In addition to optimization-based 3D shape generation, some works train a 3D generative model but leverage the embedding space of CLIP~\cite{cheng2022sdfusion,liu2023iss++,sanghi2022clip}, and some works focus on generating textures or materials for input meshes using 2D models' prior~\cite{michel2022text2mesh,wei2023taps3d,chen2023text2tex,metzer2022latent,richardson2023texture}.

\vspace{-0.8em}
\subsection{Single Image to 3D}
\vspace{-0.6em}

Before the emergence of CLIP and large-scale 2D diffusion models, people often learn 3D priors from 3D synthetic data~\cite{chang2015shapenet} or real scans~\cite{reizenstein21co3d}. Unlike 2D images, 3D data can be represented in various formats and numerous representation-specific 3D generative models have been proposed. By combing 2D image encoder and 3D generators, they generates 3D data in various representations, including 3D voxels~\cite{girdhar2016learning,wu2017marrnet, choy20163d,xie2020pix2vox++, xie2019pix2vox,yagubbayli2021legoformer}, point clouds~\cite{fan2017point,yang2018foldingnet,groueix2018papier,achlioptas2018learning,melas2023pc,zeng2022lion}, polygon meshes~\cite{kanazawa2018learning, wang2018pixel2mesh, wen2019pixel2mesh++, nash2020polygen}, and parametric models~\cite{pavlakos2019expressive, zuffi2018lions,zuffi20173d}. Recently, there has been an increasing number of work on learning to generate a 3D implicit field from a single image~\cite{xu2019disn, mescheder2019occupancy, saito2019pifu, huang2022planes, park2019deepsdf,gao2022get3d,gupta20233dgen,jang2021codenerf,muller2022autorf,wu2023multiview,mittal2022autosdf}.

As previously mentioned, several recent works leverage 2D diffusion models to perform per-shape optimization, allowing for the text-to-3D task~\cite{poole2022dreamfusion,lin2022magic3d,jain2022zero} given that diffusion models are typically conditioned on text. To enable the generation of 3D models from a single image, some works~\cite{melas2023realfusion,deng2022nerdi,metzer2022latent} utilize textual inversion~\cite{gal2022image}, to find the best-matching text embedding for the input image, which is then fed into a diffusion model. NeuralLift-360~\cite{hu2021lora} adds a CLIP loss to enforce similarity between the rendered image and the input image. 3DFuse~\cite{seo2023let} finetunes the Stable Diffusion model with LoRA layers~\cite{hu2021lora} and a sparse depth injector to ensure greater 3D consistency. A recent work Zero123~\cite{liu2023zero} finetunes the Stable Diffusion model~\cite{saharia2022photorealistic} to generate a novel view of the input image based on relative camera pose. In addition to these methods, OpenAI trains a 3D native diffusion model Point-E~\cite{nichol2022point}, which uses several million internal 3D models to generate point clouds. Very recently, they published another model Shap-E~\cite{jun2023shap} which is trained to generate parameters of implicit functions that can be used for producing textured meshes or neural radiance fields.

\vspace{-0.8em}
\subsection{Generalizable Neural Reconstruction}
\vspace{-0.6em}

Traditional NeRF-like methods~\cite{mildenhall2021nerf, wang2021neus} use a neural network to represent a single scene and require per-scene optimization. However, some approaches aim to learn priors across scenes and generalize to novel scenes. These methods typically take a few source views as input and leverage 2D networks for extracting 2D features. The pixel features are then unprojected into 3D space, and a NeRF-based rendering pipeline is applied on top of them. In this way, they can generate a 3D implicit field given a few source views in a single feed-forward pass. Among the methods, some~\cite{wang2021ibrnet, reizenstein2021common, henzler2021unsupervised, yu2021pixelnerf, yang2023contranerf, liu2022neural, kulhanek2022viewformer, trevithick2021grf, varma2022attention} directly aggregate 2D features with MLPs or transformers, while others explicitly construct the 3D feature/cost volume~\cite{chen2021mvsnerf, johari2022geonerf, zhang2022nerfusion,long2022sparseneus}, and utilize the voxel feature for decoding density and color. In addition to the density field representation, some methods such as SparseNeuS~\cite{long2022sparseneus} and VolRecon~\cite{ren2022volrecon} utilize SDF representations for geometry reconstruction. 
\vspace{-0.8em}
\section{Method}
\vspace{-0.8em}

\begin{figure}[t]
    \centering
    \includegraphics[width=\linewidth]{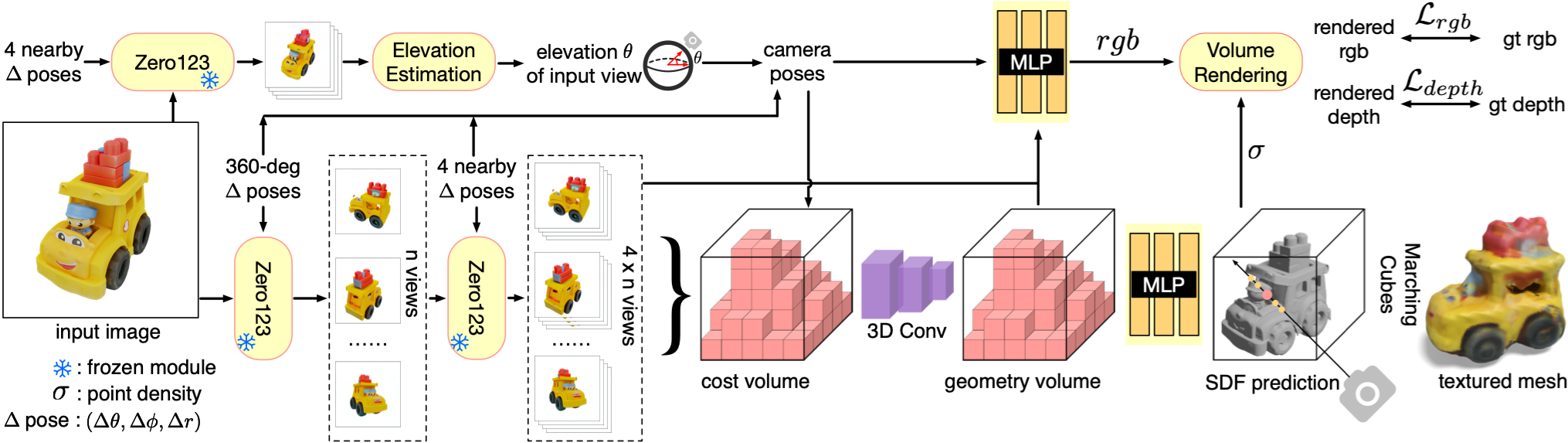}
    \vspace{-2em}
    \caption{
  Our method consists of three primary components: (a) \textbf{Multi-view synthesis}: we use a view-conditioned 2D diffusion model, Zero123~\cite{liu2023zero}, to generate multi-view images in a two-stage manner. The input of Zero123 includes a single image and a relative camera transformation, which is parameterized by the relative spherical coordinates $(\Delta \theta, \Delta \phi, \Delta r)$. (b) \textbf{Pose estimation}: we estimate the elevation angle $\theta$ of the input image based on four nearby views generated by Zero123. We then obtain the poses of the multi-view images by combining the specified relative poses with the estimated pose of the input view. (c)  \textbf{3D reconstruction}: We feed the multi-view posed images to an SDF-based generalizable neural surface reconstruction module for $360^{\circ}$ mesh reconstruction.  
    \vspace{-0.5em}}
    \label{fig:pipeline}
\end{figure}

Our overall pipeline is illustrated in Figure~\ref{fig:pipeline}. In Section~\ref{sec:Zero123}, we introduce a view-conditioned 2D diffusion model, Zero123~\cite{liu2023zero}, which is used to generate multi-view images. In Section~\ref{sec:nerf}, we show that traditional NeRF-based and SDF-based methods fail to reconstruct high-quality meshes from inconsistent multi-view predictions even given ground truth camera poses. Therefore, in Section~\ref{sec:generalizable_reconstruction}, we propose a cost volume-based neural surface reconstruction module that can be trained to handle inconsistent multi-view predictions and reconstruct a 3D mesh in a single feed-forward pass. Specifically, we build upon the SparseNeuS~\cite{long2022sparseneus} and introduce several critical training strategies to support $360^{\circ}$ mesh reconstruction. Additionally, in Section~\ref{sec:elevation_estimation}, we demonstrate the necessity of estimating the pose of the input view in Zero123's canonical space for 3D reconstruction. While the azimuth and radius can be arbitrarily specified, we propose a novel module that utilizes four nearby views generated by Zero123 to estimate the elevation of the input view.

\vspace{-0.8em}
\subsection{Zero123: View-Conditioned 2D Diffusion}
\vspace{-0.6em}
\label{sec:Zero123}

Recent 2D diffusion models~\cite{ramesh2021zero, saharia2022photorealistic, rombach2022high} have demonstrated the ability to learn a wide range of visual concepts and strong priors by training on internet-scale data. While the original diffusion models mainly focused on the task of text-to-image, recent work~\cite{zhang2023adding, hu2021lora} has shown that fine-tuning pretrained models allows us to add various conditional controls to the diffusion models and generate images based on specific conditions. Several conditions, such as canny edges, user scribbles, depth, and normal maps, have already proven effective~\cite{zhang2023adding}. 

The recent work Zero123~\cite{liu2023zero} shares a similar spirit and aims to add viewpoint condition control for the Stable Diffusion model~\cite{rombach2022high}.  Specifically, given a single RGB image of an object and a relative camera transformation, Zero123 aims to control the diffusion model to synthesize a new image under this transformed camera view. To achieve this, Zero123 fine-tunes the Stable Diffusion on paired images with their relative camera transformations, synthesized from a large-scale 3D dataset~\cite{objaverse}. During the creation of the fine-tuning dataset, Zero123 assumes that the object is centered at the origin of the coordinate system and uses a spherical camera, \ie, the camera is placed on the sphere's surface and always looks at the origin. For two camera poses $\left(\theta_1, \phi_1, r_1\right)$ and $\left(\theta_2, \phi_2, r_2\right)$, where $\theta_i$, $\phi_i$, and $r_i$ denote the polar angle, azimuth angle, and radius, their relative camera transformation is parameterized as $\left(\theta_2-\theta_1, \phi_2-\phi_1, r_2-r_1\right)$. They aim to learn a model $f$, such that $f(x_1, \theta_2-\theta_1, \phi_2-\phi_1, r_2-r_1)$ is perceptually similar to $x_2$, where $x_1$ and $x_2$ are two images of an object captured from different views. Zero123 finds that such fine-tuning enables the Stable Diffusion model to learn a generic mechanism for controlling the camera viewpoints, which extrapolates outside of the objects seen in the fine-tuning dataset. 

\vspace{-0.8em}
\subsection{Can NeRF Optimization Lift Multi-View Predictions to 3D?}
\vspace{-0.6em}
\label{sec:nerf}

\begin{figure}[t]
    \centering
    \includegraphics[width=\linewidth]{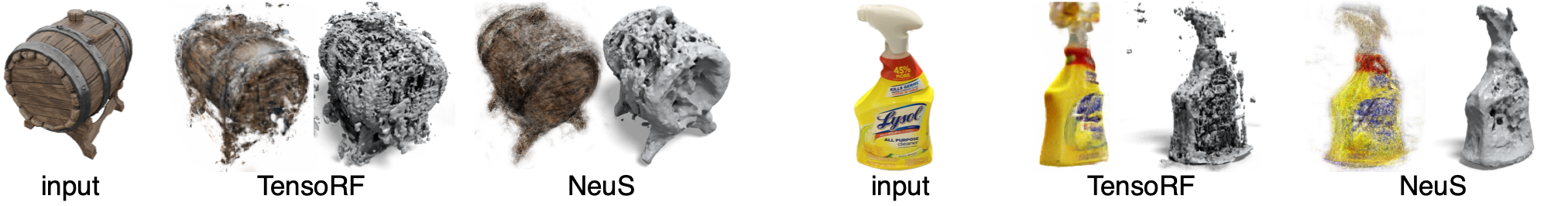}
    \vspace{-2em}
    \caption{
    NeRF-based method~\cite{mildenhall2021nerf} and SDF-based method~\cite{wang2021neus}
    fail to reconstruct high-quality meshes given multi-view images predicted by Zero123. See Figure~\ref{fig:teaser} for our reconstruction results. \vspace{-1.5em}}
    \label{fig:naive_nerf}
\end{figure}

\begin{figure}[t]
    \centering
    \includegraphics[width=\linewidth]{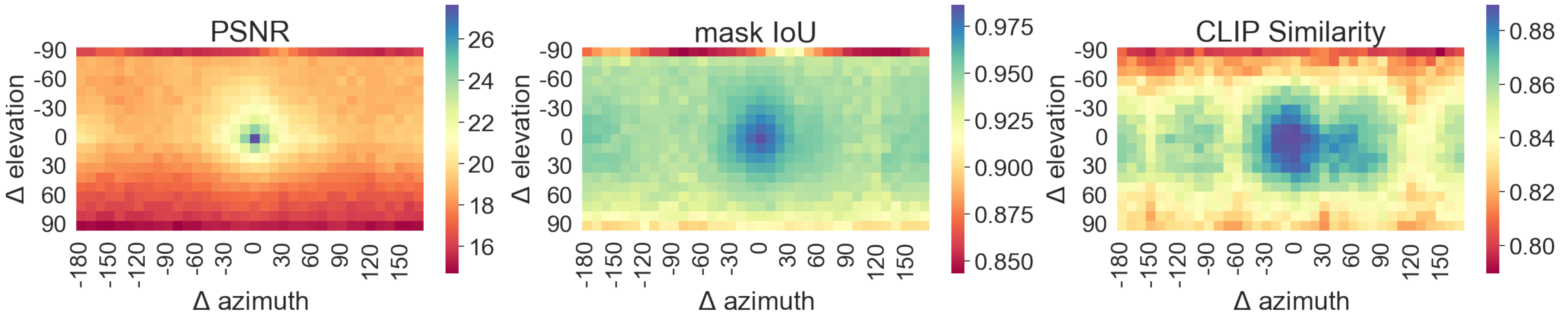}
    \vspace{-2em}
    \caption{We analyze the prediction quality of Zero123 by comparing its predictions to ground truth renderings across various view transformations. For each view transformation, we report the average PSNR, mask IoU, and CLIP similarity of 100 shapes from the Objaverse~\cite{objaverse} dataset. The prediction mask is calculated by considering foreground objects (\ie, non-white regions). Zero123 provides more accurate predictions when the view transformation is small.\vspace{-1.5em}}
    \label{fig:zero123_analysis}
    
\end{figure}

Given a single image of an object, we can utilize Zero123~\cite{liu2023zero} to generate multi-view images, but can we use traditional NeRF-based or SDF-based methods~\cite{chen2022tensorf,wang2021neus} to reconstruct high-quality 3D meshes from these predictions? We conduct a small experiment to test this hypothesis. Given a single image, we first generate 32 multi-view images using Zero123, with camera poses uniformly sampled from the sphere surface. We then feed the predictions to a NeRF-based method (TensoRF~\cite{mildenhall2021nerf}) and an SDF-based method (NeuS~\cite{wang2021neus}), which optimize density and SDF fields, respectively. However, as shown in Figure~\ref{fig:naive_nerf}, both methods fail to produce satisfactory results, generating numerous distortions and floaters. This is primarily due to the inconsistency of Zero123's predictions. In Figure~\ref{fig:zero123_analysis}, we compare Zero123's predictions with ground-truth renderings. We can see that the overall PSNR is not very high, particularly when the input relative pose is large or the target pose is at unusual locations (\eg, from the bottom or the top). However, the mask IoU (most regions are greater than 0.95) and CLIP similarity are relatively good. This suggests that Zero123 tends to generate predictions that are perceptually similar to the ground truth and have similar contours or boundaries, but the pixel-level appearance may not be exactly the same. Nevertheless, such inconsistencies between the source views are already fatal to traditional optimization-based methods. Although the original Zero123 paper proposes another method for lifting its multi-view predictions, we will demonstrate in experiments that it also fails to yield perfect results and entails time-consuming optimization.

\vspace{-0.8em}
\subsection{Neural Surface Reconstruction from Imperfect Multi-View Predictions}
\vspace{-0.6em}
\label{sec:generalizable_reconstruction}

Instead of using optimization-based approaches, we base our reconstruction module on a generalizable SDF reconstruction method SparseNeuS~\cite{long2022sparseneus}, which is essentially a variant of the MVSNeRF~\cite{chen2021mvsnerf} pipeline that combines multi-view stereo, neural scene representation, and volume rendering. As illustrated in Figure~\ref{fig:pipeline}, our reconstruction module takes multiple source images with corresponding camera poses as input and generates a textured mesh in a single feed-forward pass. In this section, we will first briefly describe the network pipeline of the module and then explain how we train the module, select the source images, and generate textured meshes. Additionally, in Section~\ref{sec:elevation_estimation}, we will discuss how we generate the camera poses for the source images.

As shown in Figure~\ref{fig:pipeline}, our reconstruction module takes $m$ posed source images as input. The module begins by extracting $m$ 2D feature maps using a 2D feature network. Next, the module builds a 3D cost volume whose contents are computed by first projecting each 3D voxel to $m$ 2D feature planes and then fetching the variance of the features across the $m$ projected 2D locations. The cost volume is then processed using a sparse 3D CNN to obtain a geometry volume that encodes the underlying geometry of the input shape. To predict the SDF at an arbitrary 3D point, an MLP network takes the 3D coordinate and its corresponding interpolated features from the geometry encoding volume as input. To predict the color of a 3D point, another MLP network takes as input the 2D features at the projected locations, interpolated features from the geometry volume, and the viewing direction of the query ray relative to the viewing direction of the source images. The network predicts the blending weights for each source view, and the color of the 3D point is predicted as the weighted sum of its projected colors. Finally, an SDF-based rendering technique is applied on top of the two MLP networks for RGB and depth rendering~\cite{wang2021neus}.

\noindent\textbf{2-Stage Source View Selection and Groundtruth-Prediction Mixed Training.} Although the original SparseNeuS~\cite{long2022sparseneus} paper only demonstrated frontal view reconstruction, we have extended it to reconstruct 360-degree meshes in a single feed-forward pass by selecting source views in a particular way and adding depth supervision during training. Specifically, our reconstruction model is trained on a 3D object dataset while freezing Zero123. We follow Zero123 to normalize the training shapes and use a spherical camera model. For each shape, we first render $n$ ground-truth RGB and depth images from $n$ camera poses uniformly placed on the sphere. For each of the $n$ views, we use Zero123 to predict four nearby views. During training, we feed all $4\times n$ predictions with ground-truth poses into the reconstruction module and randomly choose one of the $n$ ground-truth RGB images views as the target view. We call this view selection strategy as \emph{2-stage source view selection}. We supervise the training with both the ground-truth RGB and depth values. In this way, the module can learn to handle the inconsistent predictions from Zero123 and reconstruct a consistent $360^{\circ}$ mesh. We argue that our two-stage source view selection strategy is critical since uniformly choosing $n\times 4$ source views from the sphere surface would result in larger distances between the camera poses. However, cost volume-based methods~\cite{long2022sparseneus,johari2022geonerf,chen2021mvsnerf} typically rely on very close source views to find local correspondences. Furthermore, as shown in Figure~\ref{fig:zero123_analysis}, when the relative pose is small (\eg, 10 degrees apart), Zero123 can provide very accurate and consistent predictions and thus can be used to find local correspondences and infer the geometry.

During training, we use $n$ ground-truth renderings in the first stage to enable depth loss for better supervision. However, during inference, we can replace the $n$ ground-truth renderings with Zero123 predictions, as shown in Figure~\ref{fig:pipeline}, and no depth input is needed. We will show in the experiments that this groundtruth-prediction mixed training strategy is also important. To export the textured mesh, we use marching cubes~\cite{lorensen1987marching} to extract the mesh from the predicted SDF field and query the color of the mesh vertices as described in~\cite{wang2021neus}. Although our reconstruction module is trained on a 3D dataset, we find that it mainly relies on local correspondences and can generalize to unseen shapes very well.

\vspace{-0.8em}
\subsection{Camera Pose Estimation}
\vspace{-0.6em}
\label{sec:elevation_estimation}

Our reconstruction module requires camera poses for the $4\times n$ source view images. 
Note that we adopt Zero123 for image synthesis, which parameterizes cameras in a canonical spherical coordinate frame, $(\theta, \phi, r)$, where $\theta$, $\phi$ and $r$ represent the elevation, azimuth, and radius. While we can arbitrarily adjust the azimuth angle $\phi$ and the radius $r$ of all source view images simultaneously, resulting in the rotation and scaling of the reconstructed object accordingly, this parameterization requires knowing the absolute elevation angle $\theta$ of one camera to determine the relative poses of all cameras in a standard XYZ frame. More specifically, the relative poses between camera $(\theta_0, \phi_0, r_0)$ and camera $(\theta_0+\Delta\theta, \phi_0+\Delta\phi, r_0)$ vary for different $\theta_0$ even when $\Delta\theta$ and $\Delta\phi$ are the same. Because of this, changing the elevation angles of all source images together (\eg, by 30 degrees up or 30 degrees down) will lead to the distortion of the reconstructed shape (see Figure~\ref{fig:elevation} for examples).

Therefore, we propose an elevation estimation module to infer the elevation angle of the input image. First, we use Zero123 to predict four nearby views of the input image. Then we enumerate all possible elevation angles in a coarse-to-fine manner. For each elevation candidate angle, we compute the corresponding camera poses for the four images and calculate a reprojection error for this set of camera poses to measure the consistency between the images and the camera poses. The elevation angle with the smallest reprojection error is used to generate the camera poses for all $4\times n$ source views by combining the pose of the input view and the relative poses. Please refer to the supplementary for details on how we calculate the reprojection error for a set of posed images.

\vspace{-0.8em}
\section{Experiments}

\vspace{-0.8em}
\subsection{Implementation Details}
\vspace{-0.6em}

For each input image, we generate $n=8$ images by choosing camera poses uniformly placed on the sphere surface and then generate 4 local images ($10^{\circ}$ apart) for each of the 8 views, resulting in 32 source-view images for reconstruction. During training, we freeze the Zero123~\cite{liu2023zero} model and train our reconstruction module on Objaverse-LVIS~\cite{objaverse} dataset, which contains 46k 3D models in 1,156 categories. We use BlenderProc~\cite{Denninger2023} to render ground-truth RGB and depth images. For images with background, we utilize an off-the-shelf segmentation network SAM~\cite{kirillov2023segment} with bounding-box prompts for background removal. Please refer to the supplementary for more details.

\vspace{-0.8em}
\subsection{Single Image to 3D Mesh}
\vspace{-0.6em}
\begin{figure}[t]
    \centering
    \includegraphics[width=\linewidth]{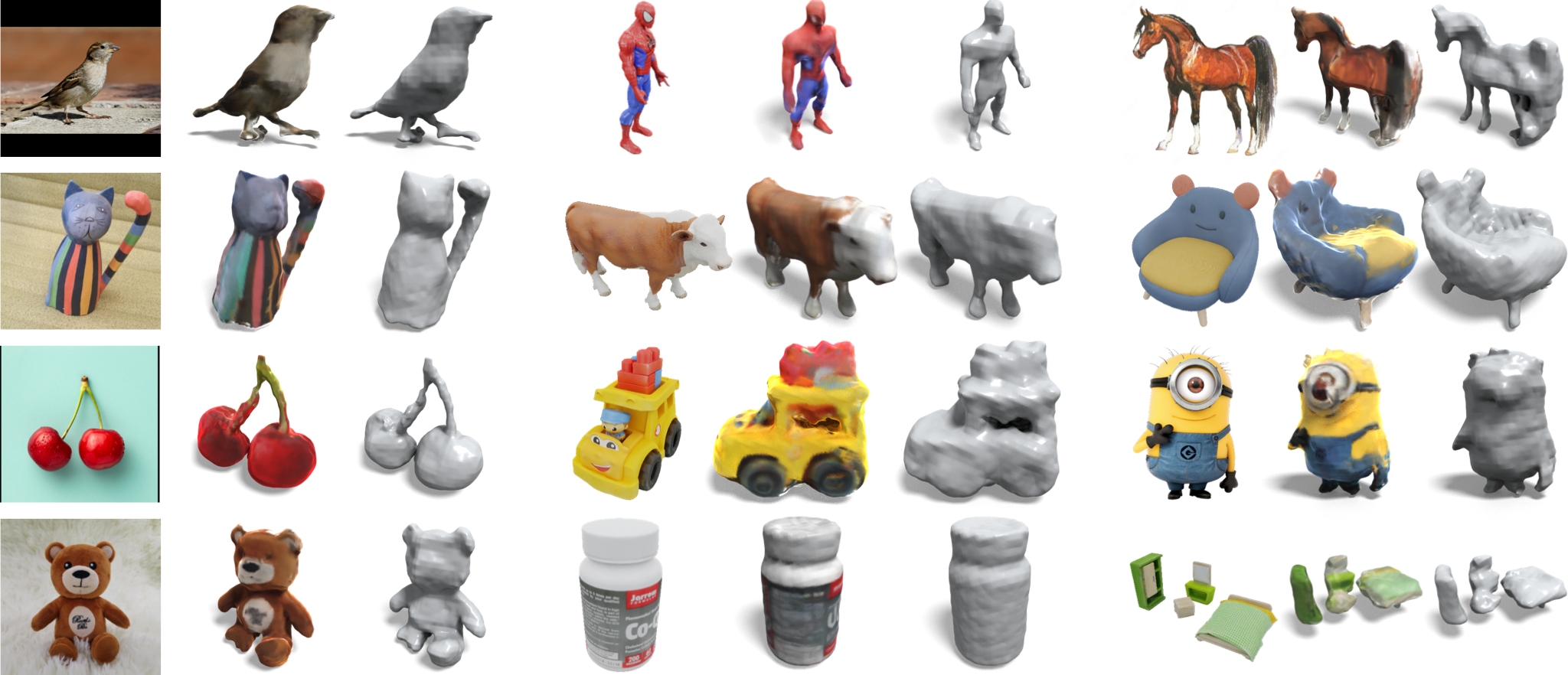}
    \vspace{-2em}
    \caption{Qualitative examples of One-2-3-45 for both synthetic and real images. Each triplet showcases an input image, a textured mesh, and a textureless mesh.\vspace{-0.5em}}
    \label{fig:qualitative}
\end{figure}

\begin{figure}[t]
    \centering
    \includegraphics[width=\linewidth]{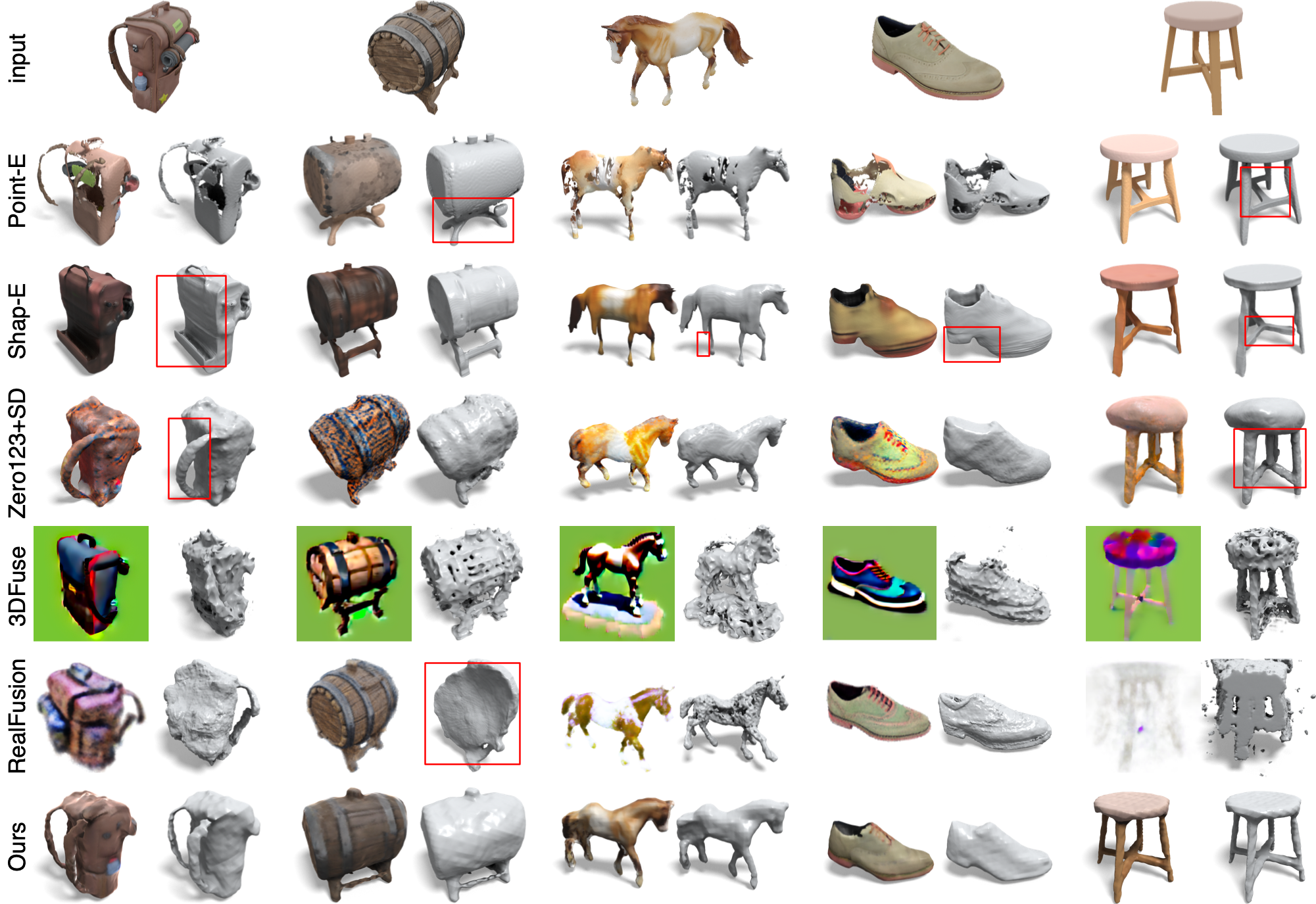}
    \vspace{-2em}
    \caption{We compare One-2-3-45 with Point-E~\cite{nichol2022point}, Shap-E~\cite{jun2023shap}, Zero123 (Stable Dreamfusion version)~\cite{liu2023zero}, 3DFuse~\cite{seo2023let}, and RealFusion~\cite{melas2023realfusion}. In each example, we present both the textured and textureless meshes. As 3DFuse~\cite{seo2023let} and RealFusion~\cite{melas2023realfusion} do not natively support the export of textured meshes, we showcase the results of volume rendering instead.   \vspace{-1.5em}}
    \label{fig:qualitative_comparison}
\end{figure}

We present qualitative examples of our method in Figures~\ref{fig:teaser} and~\ref{fig:qualitative}, illustrating its effectiveness in handling both synthetic images and real images. We also compare One-2-3-45 with existing zero-shot single image 3D reconstruction approaches, including Point-E~\cite{nichol2022point}, Shap-E~\cite{jun2023shap}, Zero123 (Stable Dreamfusion version)~\cite{liu2023zero}, 3DFuse~\cite{seo2023let}, and RealFusion~\cite{melas2023realfusion}. Among them, Point-E and Shap-E are two 3D native diffusion models released by OpenAI, which are trained on several million internal 3D data, while others are optimization-based approaches leveraging priors from Stable Diffusion~\cite{rombach2022high}. 

\begin{figure}[t]
\CenterFloatBoxes
\begin{floatrow}
\ttabbox[7.7cm]{%
\setlength{\tabcolsep}{2pt}
  \scriptsize
  \centering
  \vspace{-1em}
    \begin{tabular}{c|c|ccc|ccc|c}
    \toprule
    \multirow{2}[2]{*}{} & Prior  & \multicolumn{3}{c|}{F-Score} & \multicolumn{3}{c|}{CLIP Similarity} & \multirow{2}[2]{*}{Time} \\
          & Source & GSO   & Obj.  & avg.  & GSO   & Obj.  & avg.  &  \\
    \midrule
    Point-E~\cite{nichol2022point} & internal &    81.0   &   81.0    &   81.0 &  74.3	& 78.5 &	76.4 & 78s \\
    Shap-E~\cite{jun2023shap} & 3D data & \underline{83.4}     &  \underline{81.2}    &  \underline{82.3}   & \textbf{79.6}	& \textbf{82.1} &	\textbf{80.9} & 27s \\
    \midrule
    Zero123+SD~\cite{liu2023zero} &  \multirow{4}*{\makecell{2D \\ diffusion \\ models}}    &   75.1    &  69.9     &  72.5     & 71.0	& 72.7 &	71.9    &  $\sim$15min \\
    RealFusion~\cite{melas2023realfusion} & & 66.7     &   59.3    &  63.0     &  69.3	& 69.5 &	69.4   & $\sim$90min \\
    3DFuse~\cite{seo2023let} &   &    60.7   &   60.2    &   60.4    &   71.4	& 74.0 &	72.7 & $\sim$30min \\
    Ours  &      &   \textbf{84.0}    &  \textbf{ 83.1 }   &   \textbf{83.5}    &   \underline{76.4}	& \underline{79.7} &	\underline{78.1}    & 45s \\
    \bottomrule
    \end{tabular}%

}{%
\centering
  \caption{ Quantitative Comparison on GSO~\cite{downs2022google} and Objaverse~\cite{objaverse} datasets.} 
  \label{tab:quantitative}
}
\ffigbox[5.8cm]{%
 \includegraphics[width=0.6\linewidth]{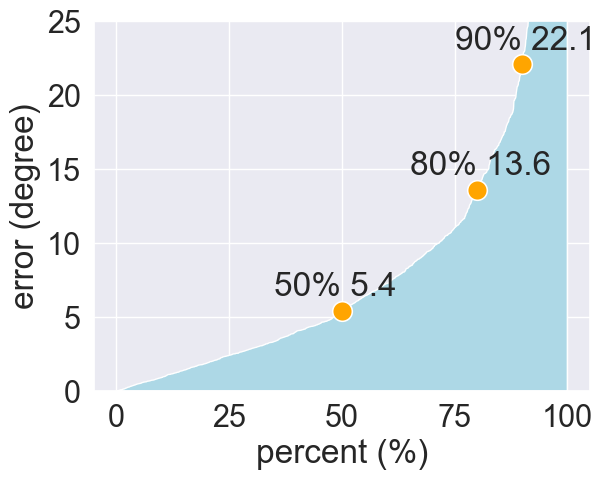}
}{%
\vspace{-1em}
  \caption{Error distribution of predicted elevations. The median and average are 5.4 and 9.7 degrees. }%
  \label{fig:elevation_error}
}
\end{floatrow}
\vspace{-1.5em}
\end{figure}

\begin{figure}[t]
    \centering
    \includegraphics[width=\linewidth]{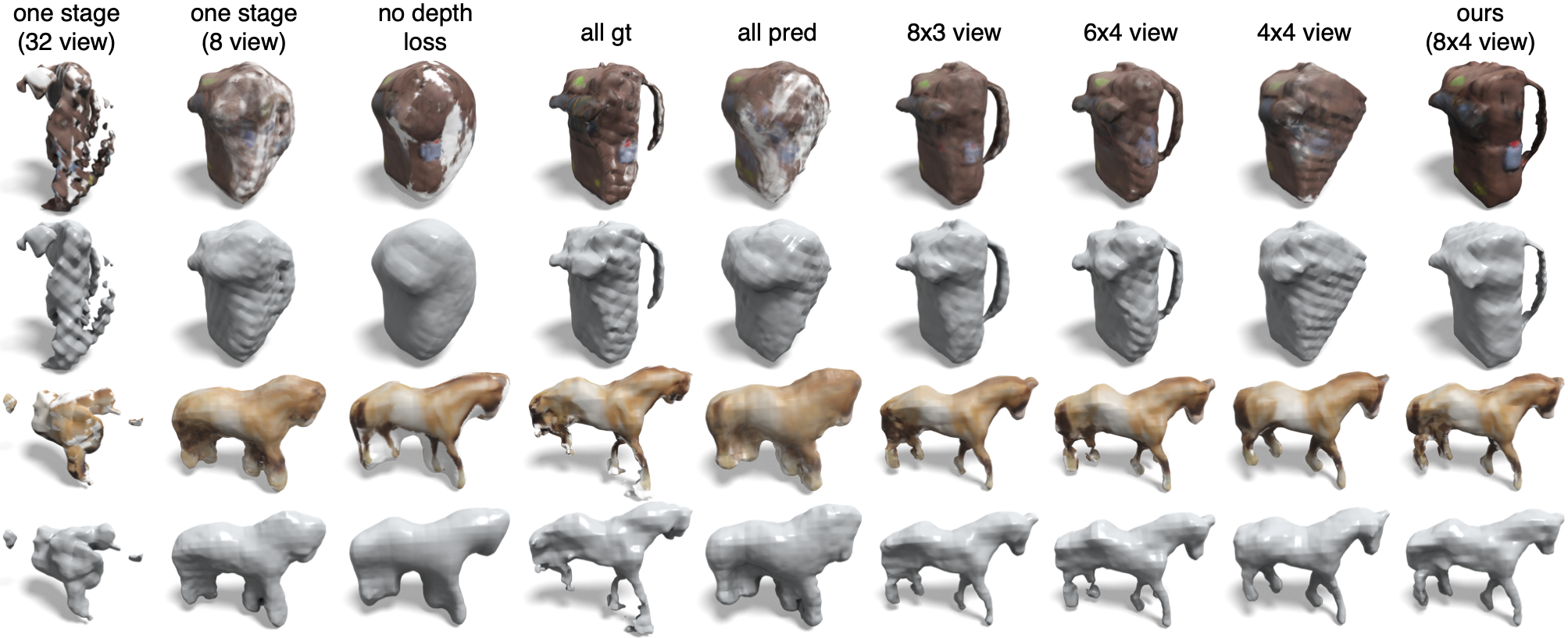}
    \vspace{-1em}
    \caption{Ablations on training strategies of the reconstruction module and the number of views. \vspace{-1.5em}}
    \vspace{-1.5em}
    \label{fig:ablation}
\end{figure}

Figure~\ref{fig:qualitative_comparison} presents the qualitative comparison. While most methods can generate plausible 3D meshes from a single image, notable differences exist among them in terms of geometry quality, adherence to the input, and overall 3D consistency. In terms of geometry quality, approaches like RealFusion~\cite{melas2023realfusion} and 3DFuse~\cite{seo2023let}, which optimize a neural radiance field, face challenges in extracting high-quality meshes. Likewise, Point-E~\cite{nichol2022point} produces a sparse point cloud as its output, resulting in numerous holes on the reconstructed meshes. In contrast, our approach utilizes an SDF presentation and favors better geometry. Regarding adherence to the input, we observe that most baseline methods struggle to preserve the similarity to the input image. Although Shap-E performs slightly better, it still produces

\begin{wrapfigure}{r}{0.28\textwidth}
  \begin{center}
  \vspace{-2em}
    \includegraphics[width=\textwidth]{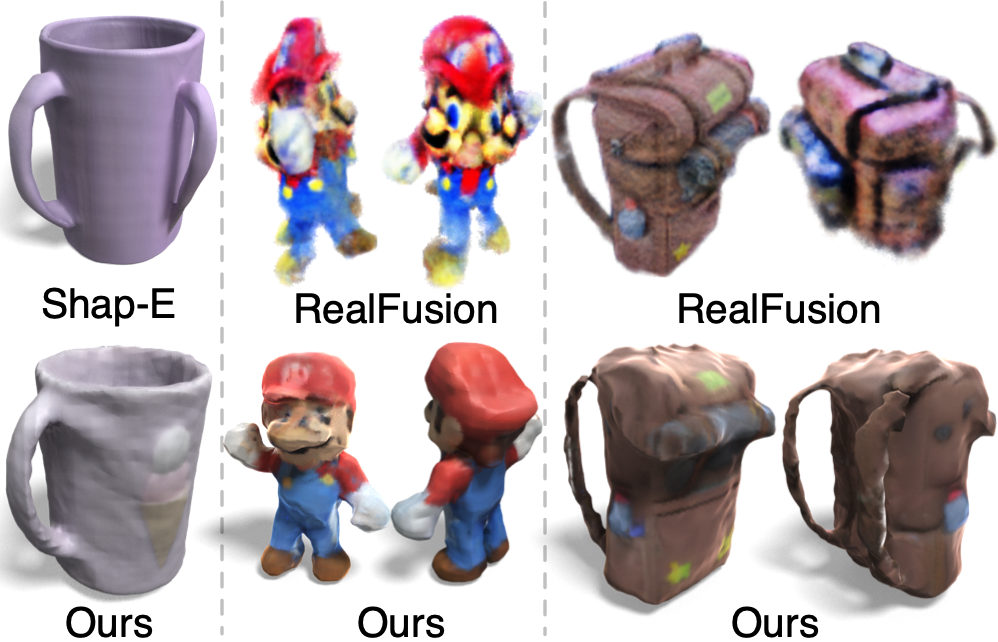}
  \end{center}
  \label{fig:two-face}
  \vspace{-2em}
\end{wrapfigure}
lots of failure cases (see the backpack without shoulder straps, distorted shoe, and stool with three legs).  In contrast, our approach leverages a powerful 2D diffusion model to directly produce high-quality multi-view images, rather than relying on 3D space hallucination. This strategy provides better adherence to the input views, alleviates the burden of the 3D reconstruction module, and yields results that are more finely attuned to the input. Furthermore, many approaches encounter challenges in achieving consistent 3D results (also known as the Janus problem~\cite{melas2023realfusion,poole2022dreamfusion}), as highlighted in the right figure (two-handle mug, multi-face Mario, and two-face backpack). One of the contributing factors to this issue is that several methods optimize each view independently, striving to make each view resemble the input. In contrast, our method capitalizes on the view-conditioned 2D diffusion model, inherently enhancing 3D consistency.

We also quantitatively compare the approaches on Objaverse~\cite{objaverse} and GoogleScannedObjects (GSO)~\cite{downs2022google} datasets. For each dataset, we randomly choose 20 shapes and render a single image per shape for evaluation. To align the predictions with the ground-truth mesh, we linearly search the scaling factor and the rotation angle, apply Iterative Closest Point (ICP) for sampled point clouds, and select the one with the most number of inliers. We follow RealFusion~\cite{melas2023realfusion} to report F-score (with a threshold of 0.05) and CLIP similarity, and the runtime on an A100 GPU. As shown in Table~\ref{tab:quantitative}, our method outperforms all baseline approaches in terms of F-Score.  As for CLIP similarity, we surpass all methods except a concurrent work Shap-E~\cite{jun2023shap}. We find that CLIP similarity is very sensitive to the color distribution and less discriminative in local geometry variations (\ie, the number of legs of a stool, the number of handles of a mug). Regarding running time, our method demonstrates a notable advantage over optimization-based approaches and performs on par with 3D native diffusion models, such as Point-E~\cite{nichol2022point} and Shap-E~\cite{jun2023shap}. Specifically, our 3D reconstruction module reconstructs a 3D mesh in approximately 5 seconds, with the remaining time primarily spent on Zero123 predictions, which take roughly 1 second per image on an A100 GPU.

\vspace{-0.8em}
\subsection{Ablation Study}
\vspace{-0.6em}

\noindent \textbf{Training strategies.} We ablate our training strategies in Figure~\ref{fig:ablation}. We found that without our 2-stage source view selection strategy, a network trained to consume 32 uniformly posed Zero123 predictions (first column) suffers from severe inconsistency among source views, causing the reconstruction module to fail completely. If we feed only 8 source views (second column) without the four nearby views, the reconstruction fails to capture local correspondence and cannot reconstruct fine-grained geometry.  Similarly, when we do not apply the depth loss during training (third column), the network fails to learn how to reconstruct fine-grained geometries. During training, we first render $n$ ground-truth renderings and then use Zero123 to predict four nearby views for each of them. If we train directly on $8\times4$ ground-truth renderings without Zero123 prediction during training (fourth column), it fails to generalize well to Zero123 predictions during inference, with many missing regions. Instead, if we replace the $n$ ground-truth renderings with $n$ Zero123 predictions during training (fifth column), the network also breaks due to the incorrect depth supervision.

\begin{figure}[t]
\CenterFloatBoxes
\begin{floatrow}

\ffigbox[6cm]{%
 \includegraphics[width=0.95\linewidth]{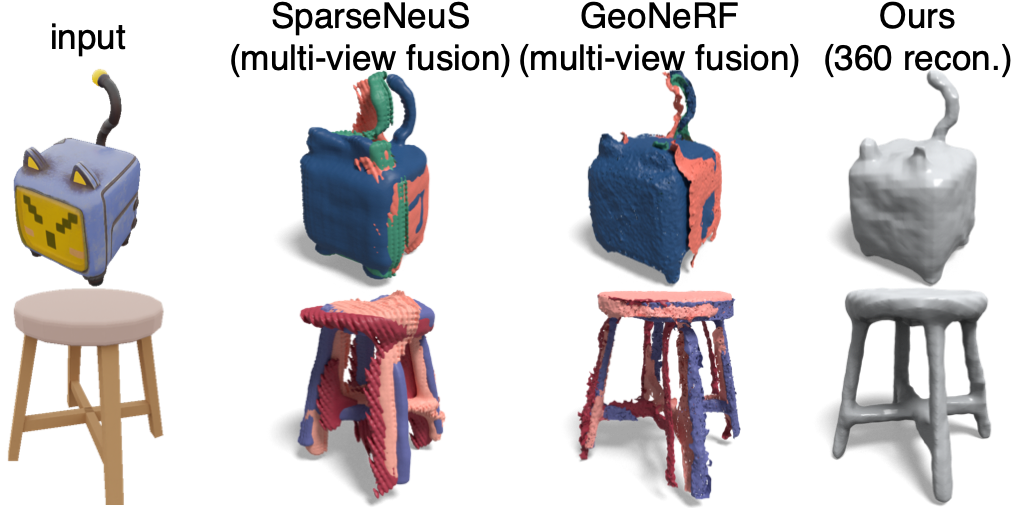}
}{%
\vspace{-1em}
  \caption{$360^{\circ}$ reconstruction vs. multi-view fusion. Meshes from different views are in different colors. \vspace{-1em}}%
  \label{fig:multi-view fusion}
}

\ffigbox[7.2cm]{%
 \includegraphics[width=0.95\linewidth]{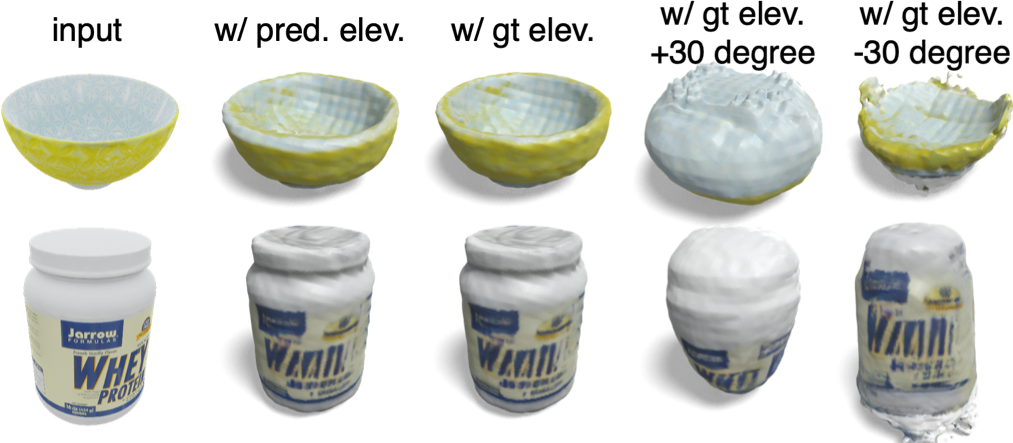}
}{%
\vspace{-1em}
  \caption{Incorrect elevations lead to distorted reconstruction. Our elevation estimation module can predict an accurate elevation of the input view. \vspace{-1em}}%
  \label{fig:elevation}
}
\end{floatrow}
\end{figure}

\noindent \textbf{Elevation estimation.} Our reconstruction module relies on accurate elevation angles of the input view. In Figure~\ref{fig:elevation}, we demonstrate the impact of providing incorrect elevation angles (\eg, altering the elevation angles of source views by $\pm 30^{\circ}$), which results in distorted reconstruction results. Instead, utilizing our predicted elevation angles can perfectly match results with ground truth elevations. We also quantitatively test our elevation estimation module by rendering 1,700 images from random camera poses. As shown in Figure~\ref{fig:elevation_error}, our elevation estimation module predicts accurate elevations.

\noindent \textbf{Number of source views.} In Figure~\ref{fig:ablation}, we also investigate the impact of varying the number of source views on 3D reconstruction. We observe that our method is not very sensitive to the number of views as long as the reconstruction module is retrained with the corresponding setting.

\noindent \textbf{$360^{\circ}$ reconstruction vs. multi-view fusion.} While our method reconstructs a $360^{\circ}$ mesh in a single pass, most existing generalizable neural reconstruction approaches~\cite{long2022sparseneus,johari2022geonerf,chen2021mvsnerf} primarily focus on frontal view reconstruction. An alternative approach is to independently infer the geometry for each view and subsequently fuse them together. However, we have observed that this strategy often struggles with multi-view fusion due to inconsistent Zero123 predictions, as illustrated in Figure~\ref{fig:multi-view fusion}.

\vspace{-0.8em}
\subsection{Text to 3D Mesh}
\vspace{-0.6em}

\begin{figure}[t]
    \centering
    \includegraphics[width=\linewidth]{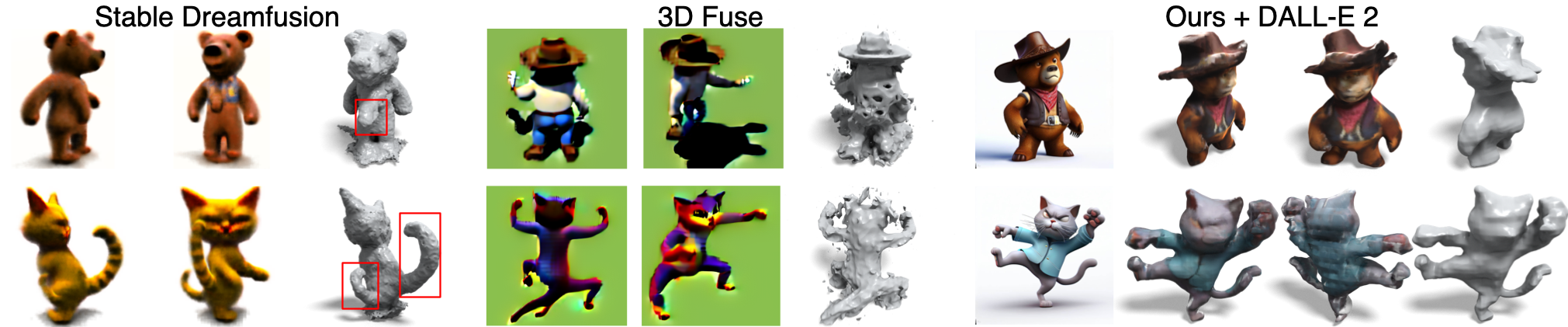}
    \vspace{-2em}
    \caption{Text to 3D. First row: ``a bear in cowboy suit.'' Second row: ``a kungfu cat.'' We utilize DALL-E 2~\cite{ramesh2022hierarchical} to generate an image conditioned on the text and then lift it to 3D. We compare our method with Stable Dreamfusion~\cite{poole2022dreamfusion} and 3DFuse~\cite{seo2023let}. For baselines, volume renderings are shown. \vspace{-2em}}
    \label{fig:text_2_3d}
\end{figure}

As shown in Figure~\ref{fig:text_2_3d}, by integrating with off-the-shelf text-to-image 2D diffusion models~\cite{rombach2022high,ramesh2022hierarchical}, our method can be naturally extended to support text-to-image-3D tasks and generate high-quality textured meshes in a short time. See supplementary for more examples.

\vspace{-0.8em}
\section{Conclusion}
\vspace{-0.8em}

In this paper, we present a novel method for reconstructing a high-quality $360^{\circ}$ mesh of any object from a single image of it. In comparison to existing zero-shot approaches, our results exhibit superior geometry, enhanced 3D consistency, and a remarkable adherence to the input image. Notably, our approach reconstructs meshes in a single forward pass without the need for time-consuming optimization, resulting in significantly reduced processing time. Furthermore, our method can be effortlessly extended to support the text-to-3D task.

\vspace{-0.5em}
\section{Appendix}
\vspace{-0.5em}

We first show more qualitative comparison in Section~\ref{sec:supp-comparison}, which is followed by a demonstration of additional examples on real-world images and the text-to-3D task in Sections~\ref{sec:supp-real} and~\ref{sec:supp-text} respectively. Furthermore, we present the details of our elevation estimation module in Section~\ref{sec:supp-estimation}, training and evaluation details in Section~\ref{sec:supp-implementation}. We finally show the failure cases and discuss the limitations in Section~\ref{sec:supp-failure}. 

\vspace{-0.5em}
\subsection{More Qualitative Comparison}
\vspace{-0.5em}
\label{sec:supp-comparison}

\begin{figure}[h]
    \centering
    \includegraphics[width=\linewidth]{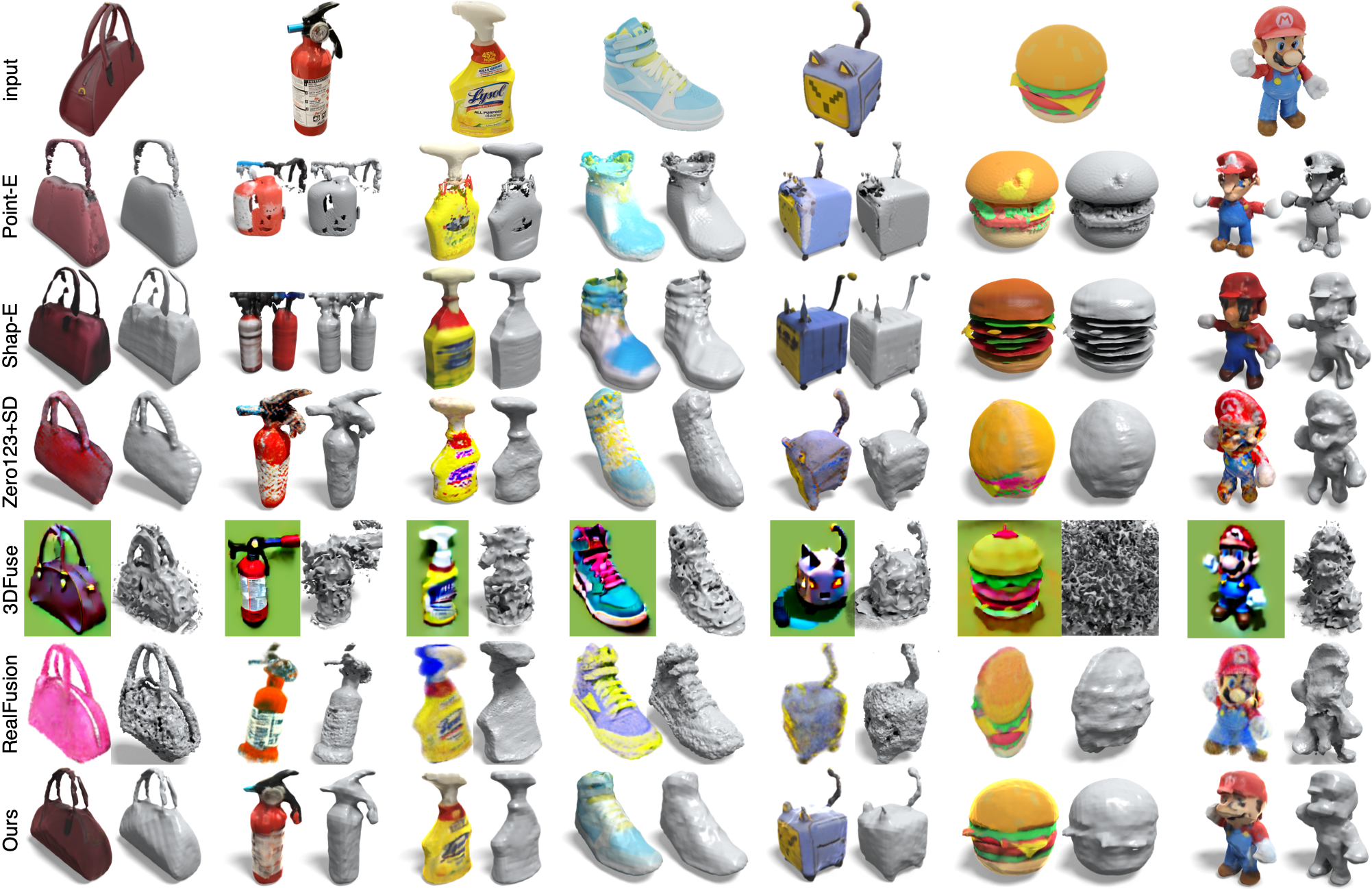}
    \caption{We compare One-2-3-45 with Point-E~\cite{nichol2022point}, Shap-E~\cite{jun2023shap}, Zero123 (Stable Dreamfusion version)~\cite{liu2023zero}, 3DFuse~\cite{seo2023let}, and RealFusion~\cite{melas2023realfusion}. In each example, we present both the textured and textureless meshes. As 3DFuse~\cite{seo2023let} and RealFusion~\cite{melas2023realfusion} do not natively support the export of textured meshes, we showcase the results of volume rendering instead.}
    \label{fig:supp_qualitative_comparsion}
\end{figure}

\begin{figure}[t]
    \centering
    \includegraphics[width=\linewidth]{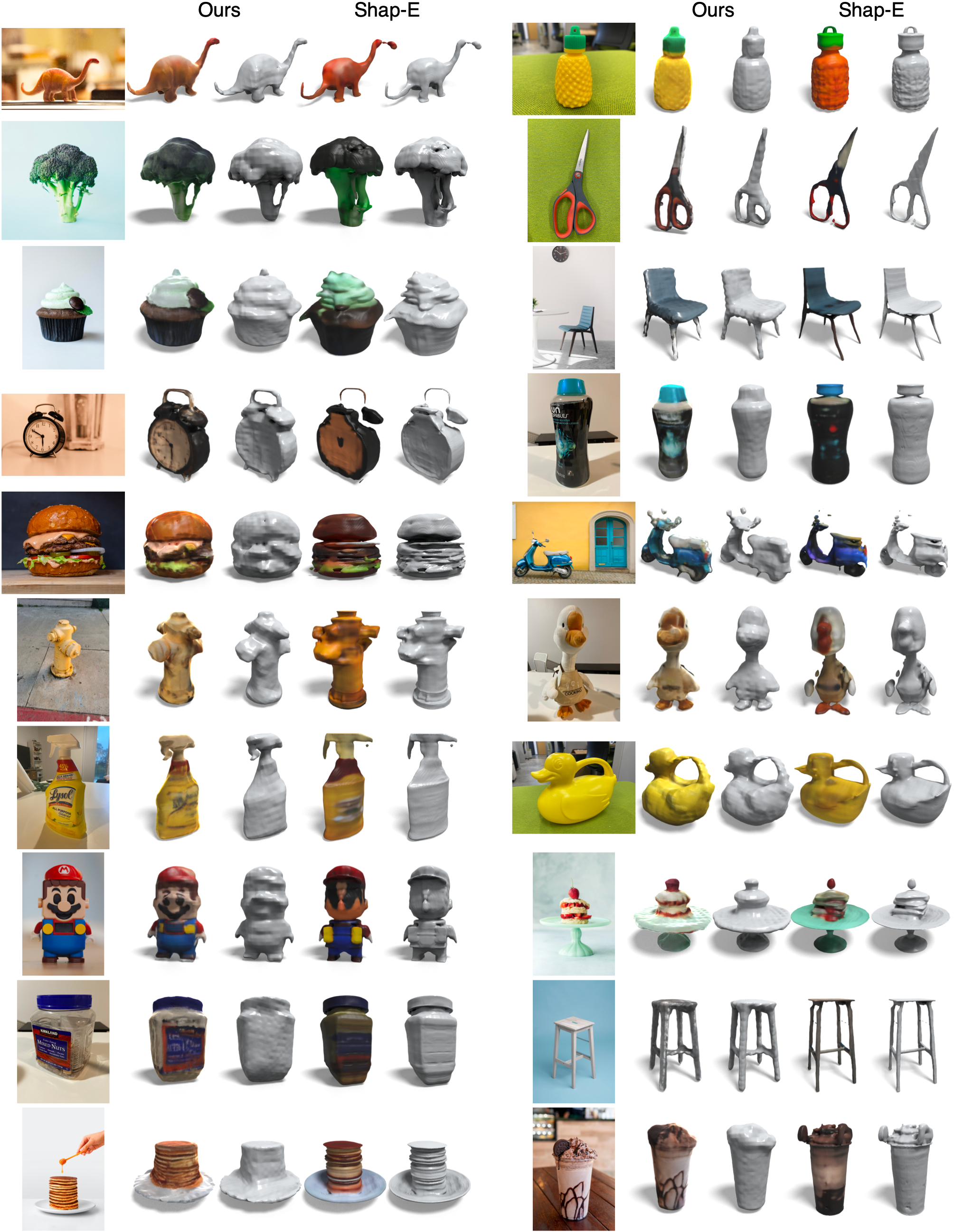}
    \caption{We compare One-2-3-45 with Shap-E~\cite{jun2023shap} on real-world images. In each example, we present the input image, generated textured and textureless meshes.}
    \vspace{-1em}
    \label{fig:supp_real}
\end{figure}

In Figure~\ref{fig:supp_qualitative_comparsion}, we demonstrate more qualitative comparison on Objaverse~\cite{objaverse} and GoogleScannedObjects (GSO)~\cite{downs2022google} datasets. Note that all test shapes are not seen during the training of our 3D reconstruction module.

\subsection{More Examples on Real-World Images}
\label{sec:supp-real}

In Figure~\ref{fig:supp_real}, we showcase more examples on real-world images and compare our method with the concurrent method Shap-E~\cite{jun2023shap}. The input images are from \url{unsplash.com} or captured by ourselves. Note that our results exhibit a closer adherence to the input image.

\subsection{More Examples on Text-to-3D}
\label{sec:supp-text}

\begin{figure}[t]
    \centering
    \includegraphics[width=\linewidth]{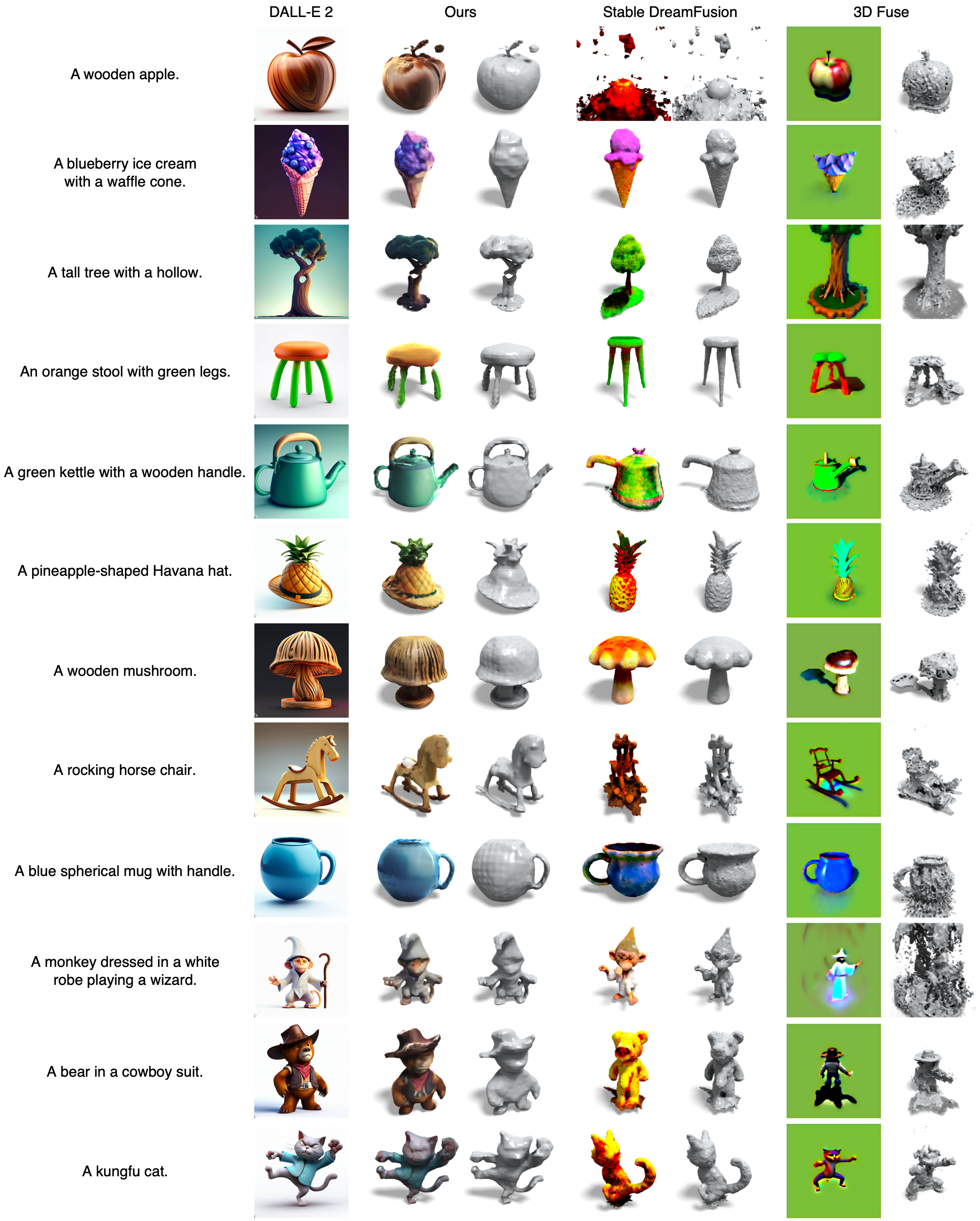}
    \caption{Text-to-3D: We compare our method against two native text-to-3D approaches Stable DreamFusion~\cite{poole2022dreamfusion} and 3DFuse~\cite{seo2023let}. To enable text-to-3D, our method first uses a pretrained text-to-image model DALL-E 2~\cite{ramesh2022hierarchical} to generate an image from input text (prompted with ``3d model, long shot''), and then uplifts the image to a 3D textured mesh. }
    \label{fig:sup_text_to_3d}
\end{figure}

In Figure~\ref{fig:sup_text_to_3d}, we present additional examples for the text-to-3D task. It is evident that existing approaches struggle to capture fine-grained details, such as a tree hollow, or achieve compositionality, as seen in examples like an orange stool with green legs, a pineapple-shaped Havana hat, or a rocking horse chair. In contrast, our method produces superior results that adhere more closely to the input text. We hypothesize that controlling such fine-grained attributes in the 3D space using existing optimization strategies is inherently challenging. However, by leveraging established 2D text-to-image diffusion models, our method becomes more effective in lifting a single 2D image to a corresponding 3D textured mesh.

\subsection{Details of Elevation Estimation}
\label{sec:supp-estimation}

To estimate the elevation angle $\theta$ of the input image, we first utilize Zero123~\cite{liu2023zero} to predict four nearby views (10 degrees apart) of the input view. With these predicted views, we proceed to enumerate all possible elevation angles and compute the re-projection error for each candidate angle. The re-projection error assesses the consistency between camera poses and image observations, akin to the bundle adjustment module employed in the Structure-from-Motion (SfM) pipeline.

Specifically, we enumerate all candidate elevation angles in a coarse-to-fine manner. In the coarse stage, we enumerate elevation angles with a 10-degree interval. Once we have determined the elevation angle $e^*$ associated with the smallest re-projection error, we proceed to the fine stage. In this stage, we enumerate elevation angle candidates ranging from $e^*-10^{\circ}$ to $e^* + 10^{\circ}$ with a 1-degree interval. This coarse-to-fine design facilitates rapid estimation, completing the elevation estimation module in under 1 second for each shape.

Given a set of four predicted nearby views, we perform feature matching to identify corresponding keypoints across each pair of images (a total of six pairs) using an off-the-shelf module LoFTR~\cite{sun2021loftr}. For each elevation angle candidate, we calculate the camera pose for the input image by employing the spherical coordinate system with a radius of 1.2 and an azimuth angle of 0. Note that the azimuth angle $\phi$ and the radius $r$ can be arbitrarily adjusted, resulting in the rotation and scaling of the reconstructed object accordingly. Subsequently, we obtain the camera poses for the four predicted views by incorporating the specified delta poses.

Once we have the four posed images, we compute the re-projection error by enumerating triplet images. For each triplet of images ($a$, $b$, $c$) sharing a set of keypoints $P$, we consider each point $p \in P$. Utilizing images $a$ and $b$, we perform triangulation to determine the 3D location of $p$. We then project the 3D point onto the third image $c$ and calculate the reprojection error, which is defined as the $l1$ distance between the reprojected 2D pixel and the estimated keypoint in image $c$. By enumerating all image triplets and their corresponding shared keypoints, we obtain the mean projection error for each elevation angle candidate.

\subsection{Details of Training and Evaluation}
\label{sec:supp-implementation}

\paragraph{Training}
We train the reconstruction module using the following loss function:
\begin{equation}
\mathcal{L} = \mathcal{L}_{rgb} + \lambda_0\mathcal{L}_{depth} + \lambda_1\mathcal{L}_{eikonal} + \lambda_2\mathcal{L}_{sparsity}
\end{equation}
where $\mathcal{L}_{rgb}$ represents the $l1$ loss between the rendered and ground truth color, weighted by the sum of accumulated weights; $\mathcal{L}_{depth}$ corresponds to the $l1$ loss between the rendered and ground truth depth; $\mathcal{L}_{eikonal}$ and $\mathcal{L}_{sparsity}$ are the Eikonal and sparsity terms, respectively, following SparseNeuS~\cite{long2022sparseneus}. We empirically set the weights as $\lambda_0 = 1$, $\lambda_1 = 0.1$, and $\lambda_2 = 0.02$. For $\lambda_2$, we adopt a linear warm-up strategy following SparseNeuS~\cite{long2022sparseneus}. To train our reconstruction module, we utilize the LVIS subset of the Objaverse~\cite{objaverse} dataset, which consists of 46k 3D models across 1,156 categories. The reconstruction module is trained for 300k iterations using two A10 GPUs, with the training process lasting approximately 6 days. It is important to note that our reconstruction module does not heavily rely on large-scale training data, as it primarily leverages local correspondence to infer the geometry, which is relatively easier to learn and generalize.

\paragraph{Evaluation}
We evaluate all baseline approaches using their official codebase. Since the approaches take only a single image as input, the predicted mesh may not have the same scale and transformation as the ground-truth mesh. To ensure a fair comparison, we employ the following process to align the predicted mesh with the ground-truth mesh. First, we align the up direction for the results generated by each approach. Next, for each generated mesh, we perform a linear search over scales and rotation angles along the up direction. After applying each pair of scale and z-rotation, we utilize the Iterative Closest Point (ICP) algorithm to align the transformed mesh to the ground-truth mesh. Finally, we select the mesh with the largest number of inliers as the final alignment. This alignment process helps us establish a consistent reference frame for evaluating the predicted meshes across different approaches.

\subsection{Failure Cases and Limitations}
\label{sec:supp-failure}

\begin{figure}[t]
    \centering
    \includegraphics[width=\linewidth]{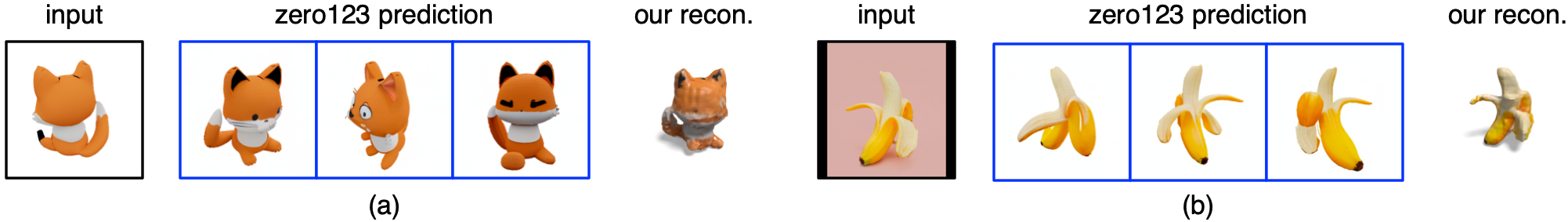}
    \caption{Failure cases. Our method relies on Zero123 to generate multi-view images, and we encounter challenges when Zero123 generates inconsistent results. (a) The input view lacks sufficient information. (b) The input view contains ambiguous or complicated structures.}
    \label{fig:sup_failure}
\end{figure}

Our method relies on Zero123 for generating multi-view images, which introduces challenges due to its occasional production of inconsistent results. In Figure~\ref{fig:sup_failure}, we present two typical cases that exemplify such inconsistencies. The first case involves an input view that lacks sufficient information, such as the back view of a fox. In this scenario, Zero123 struggles to generate consistent predictions for the invisible regions, such as the face of the fox. As a consequence, our method may encounter difficulties in accurately inferring the geometry for those regions. The second case involves an input view with ambiguous or complex structures, such as the pulp and peel of a banana. In such situations, Zero123's ability to accurately infer the underlying geometry becomes limited. As a result, our method may be affected by the inconsistent predictions generated by Zero123. It is important to acknowledge that these limitations arise from the occasional scenarios, and they can impact the performance of our method in certain cases. Addressing these challenges and refining the reliability of Zero123's predictions remain areas for further investigation and improvement.

We have also noticed slight artifacts on the back side of our generated results. As one of the first works in combining view-conditioned 2D diffusion models with generalizable multi-view reconstruction, we believe that there is still ample room for exploring more advanced reconstruction techniques and incorporating additional regularizations. By doing so, we expect to significantly mitigate the minor artifacts and further enhance results in the future.

\subsection{Acknowledgements} 
We would like to thank the following sketchfab users for the models used for the demo images in this paper: dimaponomar2019 (backpack), danielpeng (bag), pmlzbt233 (wooden barrel), felixyadomi (cactus), avianinda (burger), shedmon (robocat), ie-niels (stool), phucn (armchair), techCIR (mug), sabriny (fox). All models are CC-By licensed.

{
\bibliographystyle{ieee_fullname}
\bibliography{egbib}

\begin{thebibliography}{10}\itemsep=-1pt

\bibitem{achlioptas2018learning}
Panos Achlioptas, Olga Diamanti, Ioannis Mitliagkas, and Leonidas Guibas.
\newblock Learning representations and generative models for 3d point clouds.
\newblock In {\em International conference on machine learning}, pages 40--49.
  PMLR, 2018.

\bibitem{aneja2022clipface}
Shivangi Aneja, Justus Thies, Angela Dai, and Matthias Nie{\ss}ner.
\newblock Clipface: Text-guided editing of textured 3d morphable models.
\newblock {\em arXiv preprint arXiv:2212.01406}, 2022.

\bibitem{canfes2023text}
Zehranaz Canfes, M~Furkan Atasoy, Alara Dirik, and Pinar Yanardag.
\newblock Text and image guided 3d avatar generation and manipulation.
\newblock In {\em Proceedings of the IEEE/CVF Winter Conference on Applications
  of Computer Vision}, pages 4421--4431, 2023.

\bibitem{chang2015shapenet}
Angel~X Chang, Thomas Funkhouser, Leonidas Guibas, Pat Hanrahan, Qixing Huang,
  Zimo Li, Silvio Savarese, Manolis Savva, Shuran Song, Hao Su, et~al.
\newblock Shapenet: An information-rich 3d model repository.
\newblock {\em arXiv preprint arXiv:1512.03012}, 2015.

\bibitem{chen2022tensorf}
Anpei Chen, Zexiang Xu, Andreas Geiger, Jingyi Yu, and Hao Su.
\newblock Tensorf: Tensorial radiance fields.
\newblock In {\em Computer Vision--ECCV 2022: 17th European Conference, Tel
  Aviv, Israel, October 23--27, 2022, Proceedings, Part XXXII}, pages 333--350.
  Springer, 2022.

\bibitem{chen2021mvsnerf}
Anpei Chen, Zexiang Xu, Fuqiang Zhao, Xiaoshuai Zhang, Fanbo Xiang, Jingyi Yu,
  and Hao Su.
\newblock Mvsnerf: Fast generalizable radiance field reconstruction from
  multi-view stereo.
\newblock In {\em Proceedings of the IEEE/CVF International Conference on
  Computer Vision}, pages 14124--14133, 2021.

\bibitem{chen2023text2tex}
Dave~Zhenyu Chen, Yawar Siddiqui, Hsin-Ying Lee, Sergey Tulyakov, and Matthias
  Nie{\ss}ner.
\newblock Text2tex: Text-driven texture synthesis via diffusion models.
\newblock {\em arXiv preprint arXiv:2303.11396}, 2023.

\bibitem{cheng2022sdfusion}
Yen-Chi Cheng, Hsin-Ying Lee, Sergey Tulyakov, Alexander Schwing, and Liangyan
  Gui.
\newblock Sdfusion: Multimodal 3d shape completion, reconstruction, and
  generation.
\newblock {\em arXiv preprint arXiv:2212.04493}, 2022.

\bibitem{chiu2009automatic}
Han-Pang Chiu, Leslie~Pack Kaelbling, and Tom{\'a}s Lozano-P{\'e}rez.
\newblock Automatic class-specific 3d reconstruction from a single image.
\newblock {\em CSAIL}, pages 1--9, 2009.

\bibitem{choy20163d}
Christopher~B Choy, Danfei Xu, JunYoung Gwak, Kevin Chen, and Silvio Savarese.
\newblock 3d-r2n2: A unified approach for single and multi-view 3d object
  reconstruction.
\newblock In {\em Computer Vision--ECCV 2016: 14th European Conference,
  Amsterdam, The Netherlands, October 11-14, 2016, Proceedings, Part VIII 14},
  pages 628--644. Springer, 2016.

\bibitem{objaverse}
Matt Deitke, Dustin Schwenk, Jordi Salvador, Luca Weihs, Oscar Michel, Eli
  VanderBilt, Ludwig Schmidt, Kiana Ehsani, Aniruddha Kembhavi, and Ali
  Farhadi.
\newblock Objaverse: A universe of annotated 3d objects.
\newblock {\em arXiv preprint arXiv:2212.08051}, 2022.

\bibitem{deng2022nerdi}
Congyue Deng, Chiyu Jiang, Charles~R Qi, Xinchen Yan, Yin Zhou, Leonidas
  Guibas, Dragomir Anguelov, et~al.
\newblock Nerdi: Single-view nerf synthesis with language-guided diffusion as
  general image priors.
\newblock {\em arXiv preprint arXiv:2212.03267}, 2022.

\bibitem{Denninger2023}
Maximilian Denninger, Dominik Winkelbauer, Martin Sundermeyer, Wout Boerdijk,
  Markus Knauer, Klaus~H. Strobl, Matthias Humt, and Rudolph Triebel.
\newblock Blenderproc2: A procedural pipeline for photorealistic rendering.
\newblock {\em Journal of Open Source Software}, 8(82):4901, 2023.

\bibitem{downs2022google}
Laura Downs, Anthony Francis, Nate Koenig, Brandon Kinman, Ryan Hickman, Krista
  Reymann, Thomas~B McHugh, and Vincent Vanhoucke.
\newblock Google scanned objects: A high-quality dataset of 3d scanned
  household items.
\newblock In {\em 2022 International Conference on Robotics and Automation
  (ICRA)}, pages 2553--2560. IEEE, 2022.

\bibitem{fan2017point}
Haoqiang Fan, Hao Su, and Leonidas~J Guibas.
\newblock A point set generation network for 3d object reconstruction from a
  single image.
\newblock In {\em Proceedings of the IEEE conference on computer vision and
  pattern recognition}, pages 605--613, 2017.

\bibitem{gal2022image}
Rinon Gal, Yuval Alaluf, Yuval Atzmon, Or Patashnik, Amit~H Bermano, Gal
  Chechik, and Daniel Cohen-Or.
\newblock An image is worth one word: Personalizing text-to-image generation
  using textual inversion.
\newblock {\em arXiv preprint arXiv:2208.01618}, 2022.

\bibitem{gao2022get3d}
Jun Gao, Tianchang Shen, Zian Wang, Wenzheng Chen, Kangxue Yin, Daiqing Li, Or
  Litany, Zan Gojcic, and Sanja Fidler.
\newblock Get3d: A generative model of high quality 3d textured shapes learned
  from images.
\newblock {\em Advances In Neural Information Processing Systems},
  35:31841--31854, 2022.

\bibitem{girdhar2016learning}
Rohit Girdhar, David~F Fouhey, Mikel Rodriguez, and Abhinav Gupta.
\newblock Learning a predictable and generative vector representation for
  objects.
\newblock In {\em Computer Vision--ECCV 2016: 14th European Conference,
  Amsterdam, The Netherlands, October 11-14, 2016, Proceedings, Part VI 14},
  pages 484--499. Springer, 2016.

\bibitem{groueix2018papier}
Thibault Groueix, Matthew Fisher, Vladimir~G Kim, Bryan~C Russell, and Mathieu
  Aubry.
\newblock A papier-m{\^a}ch{\'e} approach to learning 3d surface generation.
\newblock In {\em Proceedings of the IEEE conference on computer vision and
  pattern recognition}, pages 216--224, 2018.

\bibitem{gupta20233dgen}
Anchit Gupta, Wenhan Xiong, Yixin Nie, Ian Jones, and Barlas O{\u{g}}uz.
\newblock 3dgen: Triplane latent diffusion for textured mesh generation.
\newblock {\em arXiv preprint arXiv:2303.05371}, 2023.

\bibitem{henzler2021unsupervised}
Philipp Henzler, Jeremy Reizenstein, Patrick Labatut, Roman Shapovalov, Tobias
  Ritschel, Andrea Vedaldi, and David Novotny.
\newblock Unsupervised learning of 3d object categories from videos in the
  wild.
\newblock In {\em Proceedings of the IEEE/CVF Conference on Computer Vision and
  Pattern Recognition}, pages 4700--4709, 2021.

\bibitem{hong2022avatarclip}
Fangzhou Hong, Mingyuan Zhang, Liang Pan, Zhongang Cai, Lei Yang, and Ziwei
  Liu.
\newblock Avatarclip: Zero-shot text-driven generation and animation of 3d
  avatars.
\newblock {\em arXiv preprint arXiv:2205.08535}, 2022.

\bibitem{hu2021lora}
Edward~J Hu, Yelong Shen, Phillip Wallis, Zeyuan Allen-Zhu, Yuanzhi Li, Shean
  Wang, Lu Wang, and Weizhu Chen.
\newblock Lora: Low-rank adaptation of large language models.
\newblock {\em arXiv preprint arXiv:2106.09685}, 2021.

\bibitem{huang2022planes}
Zixuan Huang, Stefan Stojanov, Anh Thai, Varun Jampani, and James~M Rehg.
\newblock Planes vs. chairs: Category-guided 3d shape learning without any 3d
  cues.
\newblock In {\em Computer Vision--ECCV 2022: 17th European Conference, Tel
  Aviv, Israel, October 23--27, 2022, Proceedings, Part I}, pages 727--744.
  Springer, 2022.

\bibitem{jain2022zero}
Ajay Jain, Ben Mildenhall, Jonathan~T Barron, Pieter Abbeel, and Ben Poole.
\newblock Zero-shot text-guided object generation with dream fields.
\newblock In {\em Proceedings of the IEEE/CVF Conference on Computer Vision and
  Pattern Recognition}, pages 867--876, 2022.

\bibitem{jang2021codenerf}
Wonbong Jang and Lourdes Agapito.
\newblock Codenerf: Disentangled neural radiance fields for object categories.
\newblock In {\em Proceedings of the IEEE/CVF International Conference on
  Computer Vision}, pages 12949--12958, 2021.

\bibitem{jetchev2021clipmatrix}
Nikolay Jetchev.
\newblock Clipmatrix: Text-controlled creation of 3d textured meshes.
\newblock {\em arXiv preprint arXiv:2109.12922}, 2021.

\bibitem{johari2022geonerf}
Mohammad~Mahdi Johari, Yann Lepoittevin, and Fran{\c{c}}ois Fleuret.
\newblock Geonerf: Generalizing nerf with geometry priors.
\newblock In {\em Proceedings of the IEEE/CVF Conference on Computer Vision and
  Pattern Recognition}, pages 18365--18375, 2022.

\bibitem{jun2023shap}
Heewoo Jun and Alex Nichol.
\newblock Shap-e: Generating conditional 3d implicit functions.
\newblock {\em arXiv preprint arXiv:2305.02463}, 2023.

\bibitem{kanazawa2018learning}
Angjoo Kanazawa, Shubham Tulsiani, Alexei~A Efros, and Jitendra Malik.
\newblock Learning category-specific mesh reconstruction from image
  collectionsgirdhar2016learning.
\newblock In {\em Proceedings of the European Conference on Computer Vision
  (ECCV)}, pages 371--386, 2018.

\bibitem{khalid2022text}
Nasir Khalid, Tianhao Xie, Eugene Belilovsky, and Tiberiu Popa.
\newblock Text to mesh without 3d supervision using limit subdivision.
\newblock {\em arXiv preprint arXiv:2203.13333}, 2022.

\bibitem{kirillov2023segment}
Alexander Kirillov, Eric Mintun, Nikhila Ravi, Hanzi Mao, Chloe Rolland, Laura
  Gustafson, Tete Xiao, Spencer Whitehead, Alexander~C Berg, Wan-Yen Lo, et~al.
\newblock Segment anything.
\newblock {\em arXiv preprint arXiv:2304.02643}, 2023.

\bibitem{kulhanek2022viewformer}
Jon{\'a}{\v{s}} Kulh{\'a}nek, Erik Derner, Torsten Sattler, and Robert
  Babu{\v{s}}ka.
\newblock Viewformer: Nerf-free neural rendering from few images using
  transformers.
\newblock In {\em Computer Vision--ECCV 2022: 17th European Conference, Tel
  Aviv, Israel, October 23--27, 2022, Proceedings, Part XV}, pages 198--216.
  Springer, 2022.

\bibitem{lee2022understanding}
Han-Hung Lee and Angel~X Chang.
\newblock Understanding pure clip guidance for voxel grid nerf models.
\newblock {\em arXiv preprint arXiv:2209.15172}, 2022.

\bibitem{lin2022magic3d}
Chen-Hsuan Lin, Jun Gao, Luming Tang, Towaki Takikawa, Xiaohui Zeng, Xun Huang,
  Karsten Kreis, Sanja Fidler, Ming-Yu Liu, and Tsung-Yi Lin.
\newblock Magic3d: High-resolution text-to-3d content creation.
\newblock {\em arXiv preprint arXiv:2211.10440}, 2022.

\bibitem{liu2023zero}
Ruoshi Liu, Rundi Wu, Basile Van~Hoorick, Pavel Tokmakov, Sergey Zakharov, and
  Carl Vondrick.
\newblock Zero-1-to-3: Zero-shot one image to 3d object.
\newblock {\em arXiv preprint arXiv:2303.11328}, 2023.

\bibitem{liu2022neural}
Yuan Liu, Sida Peng, Lingjie Liu, Qianqian Wang, Peng Wang, Christian Theobalt,
  Xiaowei Zhou, and Wenping Wang.
\newblock Neural rays for occlusion-aware image-based rendering.
\newblock In {\em Proceedings of the IEEE/CVF Conference on Computer Vision and
  Pattern Recognition}, pages 7824--7833, 2022.

\bibitem{liu2022iss}
Zhengzhe Liu, Peng Dai, Ruihui Li, Xiaojuan Qi, and Chi-Wing Fu.
\newblock Iss: Image as stetting stone for text-guided 3d shape generation.
\newblock {\em arXiv preprint arXiv:2209.04145}, 2022.

\bibitem{liu2023iss++}
Zhengzhe Liu, Peng Dai, Ruihui Li, Xiaojuan Qi, and Chi-Wing Fu.
\newblock Iss++: Image as stepping stone for text-guided 3d shape generation.
\newblock {\em arXiv preprint arXiv:2303.15181}, 2023.

\bibitem{long2022sparseneus}
Xiaoxiao Long, Cheng Lin, Peng Wang, Taku Komura, and Wenping Wang.
\newblock Sparseneus: Fast generalizable neural surface reconstruction from
  sparse views.
\newblock In {\em Computer Vision--ECCV 2022: 17th European Conference, Tel
  Aviv, Israel, October 23--27, 2022, Proceedings, Part XXXII}, pages 210--227.
  Springer, 2022.

\bibitem{lorensen1987marching}
William~E Lorensen and Harvey~E Cline.
\newblock Marching cubes: A high resolution 3d surface construction algorithm.
\newblock {\em ACM siggraph computer graphics}, 21(4):163--169, 1987.

\bibitem{mees2019self}
Oier Mees, Maxim Tatarchenko, Thomas Brox, and Wolfram Burgard.
\newblock Self-supervised 3d shape and viewpoint estimation from single images
  for robotics.
\newblock In {\em 2019 IEEE/RSJ International Conference on Intelligent Robots
  and Systems (IROS)}, pages 6083--6089. IEEE, 2019.

\bibitem{melas2023realfusion}
Luke Melas-Kyriazi, Christian Rupprecht, Iro Laina, and Andrea Vedaldi.
\newblock Realfusion: 360 $\{$$\backslash$deg$\}$ reconstruction of any object
  from a single image.
\newblock {\em arXiv preprint arXiv:2302.10663}, 2023.

\bibitem{melas2023pc}
Luke Melas-Kyriazi, Christian Rupprecht, and Andrea Vedaldi.
\newblock $ pc^{} 2$: Projection—conditioned point cloud diffusion for
  single-image 3d reconstruction.
\newblock {\em arXiv preprint arXiv:2302.10668}, 2023.

\bibitem{mescheder2019occupancy}
Lars Mescheder, Michael Oechsle, Michael Niemeyer, Sebastian Nowozin, and
  Andreas Geiger.
\newblock Occupancy networks: Learning 3d reconstruction in function space.
\newblock In {\em Proceedings of the IEEE/CVF conference on computer vision and
  pattern recognition}, pages 4460--4470, 2019.

\bibitem{metzer2022latent}
Gal Metzer, Elad Richardson, Or Patashnik, Raja Giryes, and Daniel Cohen-Or.
\newblock Latent-nerf for shape-guided generation of 3d shapes and textures.
\newblock {\em arXiv preprint arXiv:2211.07600}, 2022.

\bibitem{michel2022text2mesh}
Oscar Michel, Roi Bar-On, Richard Liu, Sagie Benaim, and Rana Hanocka.
\newblock Text2mesh: Text-driven neural stylization for meshes.
\newblock In {\em Proceedings of the IEEE/CVF Conference on Computer Vision and
  Pattern Recognition}, pages 13492--13502, 2022.

\bibitem{mildenhall2021nerf}
Ben Mildenhall, Pratul~P Srinivasan, Matthew Tancik, Jonathan~T Barron, Ravi
  Ramamoorthi, and Ren Ng.
\newblock Nerf: Representing scenes as neural radiance fields for view
  synthesis.
\newblock {\em Communications of the ACM}, 65(1):99--106, 2021.

\bibitem{mittal2022autosdf}
Paritosh Mittal, Yen-Chi Cheng, Maneesh Singh, and Shubham Tulsiani.
\newblock Autosdf: Shape priors for 3d completion, reconstruction and
  generation.
\newblock In {\em Proceedings of the IEEE/CVF Conference on Computer Vision and
  Pattern Recognition}, pages 306--315, 2022.

\bibitem{muller2022autorf}
Norman M{\"u}ller, Andrea Simonelli, Lorenzo Porzi, Samuel~Rota Bul{\`o},
  Matthias Nie{\ss}ner, and Peter Kontschieder.
\newblock Autorf: Learning 3d object radiance fields from single view
  observations.
\newblock In {\em Proceedings of the IEEE/CVF Conference on Computer Vision and
  Pattern Recognition}, pages 3971--3980, 2022.

\bibitem{nash2020polygen}
Charlie Nash, Yaroslav Ganin, SM~Ali Eslami, and Peter Battaglia.
\newblock Polygen: An autoregressive generative model of 3d meshes.
\newblock In {\em International conference on machine learning}, pages
  7220--7229. PMLR, 2020.

\bibitem{nichol2022point}
Alex Nichol, Heewoo Jun, Prafulla Dhariwal, Pamela Mishkin, and Mark Chen.
\newblock Point-e: A system for generating 3d point clouds from complex
  prompts.
\newblock {\em arXiv preprint arXiv:2212.08751}, 2022.

\bibitem{park2019deepsdf}
Jeong~Joon Park, Peter Florence, Julian Straub, Richard Newcombe, and Steven
  Lovegrove.
\newblock Deepsdf: Learning continuous signed distance functions for shape
  representation.
\newblock In {\em Proceedings of the IEEE/CVF conference on computer vision and
  pattern recognition}, pages 165--174, 2019.

\bibitem{pavlakos2019expressive}
Georgios Pavlakos, Vasileios Choutas, Nima Ghorbani, Timo Bolkart, Ahmed~AA
  Osman, Dimitrios Tzionas, and Michael~J Black.
\newblock Expressive body capture: 3d hands, face, and body from a single
  image.
\newblock In {\em Proceedings of the IEEE/CVF conference on computer vision and
  pattern recognition}, pages 10975--10985, 2019.

\bibitem{poole2022dreamfusion}
Ben Poole, Ajay Jain, Jonathan~T Barron, and Ben Mildenhall.
\newblock Dreamfusion: Text-to-3d using 2d diffusion.
\newblock {\em arXiv preprint arXiv:2209.14988}, 2022.

\bibitem{radford2021learning}
Alec Radford, Jong~Wook Kim, Chris Hallacy, Aditya Ramesh, Gabriel Goh,
  Sandhini Agarwal, Girish Sastry, Amanda Askell, Pamela Mishkin, Jack Clark,
  et~al.
\newblock Learning transferable visual models from natural language
  supervision.
\newblock In {\em International Conference on Machine Learning}, pages
  8748--8763. PMLR, 2021.

\bibitem{raj2023dreambooth3d}
Amit Raj, Srinivas Kaza, Ben Poole, Michael Niemeyer, Nataniel Ruiz, Ben
  Mildenhall, Shiran Zada, Kfir Aberman, Michael Rubinstein, Jonathan Barron,
  et~al.
\newblock Dreambooth3d: Subject-driven text-to-3d generation.
\newblock {\em arXiv preprint arXiv:2303.13508}, 2023.

\bibitem{ramesh2022hierarchical}
Aditya Ramesh, Prafulla Dhariwal, Alex Nichol, Casey Chu, and Mark Chen.
\newblock Hierarchical text-conditional image generation with clip latents.
\newblock {\em arXiv preprint arXiv:2204.06125}, 2022.

\bibitem{ramesh2021zero}
Aditya Ramesh, Mikhail Pavlov, Gabriel Goh, Scott Gray, Chelsea Voss, Alec
  Radford, Mark Chen, and Ilya Sutskever.
\newblock Zero-shot text-to-image generation.
\newblock In {\em International Conference on Machine Learning}, pages
  8821--8831. PMLR, 2021.

\bibitem{reizenstein21co3d}
Jeremy Reizenstein, Roman Shapovalov, Philipp Henzler, Luca Sbordone, Patrick
  Labatut, and David Novotny.
\newblock Common objects in 3d: Large-scale learning and evaluation of
  real-life 3d category reconstruction.
\newblock In {\em International Conference on Computer Vision}, 2021.

\bibitem{reizenstein2021common}
Jeremy Reizenstein, Roman Shapovalov, Philipp Henzler, Luca Sbordone, Patrick
  Labatut, and David Novotny.
\newblock Common objects in 3d: Large-scale learning and evaluation of
  real-life 3d category reconstruction.
\newblock In {\em Proceedings of the IEEE/CVF International Conference on
  Computer Vision}, pages 10901--10911, 2021.

\bibitem{ren2022volrecon}
Yufan Ren, Fangjinhua Wang, Tong Zhang, Marc Pollefeys, and Sabine
  S{\"u}sstrunk.
\newblock Volrecon: Volume rendering of signed ray distance functions for
  generalizable multi-view reconstruction.
\newblock {\em arXiv preprint arXiv:2212.08067}, 2022.

\bibitem{richardson2023texture}
Elad Richardson, Gal Metzer, Yuval Alaluf, Raja Giryes, and Daniel Cohen-Or.
\newblock Texture: Text-guided texturing of 3d shapes.
\newblock {\em arXiv preprint arXiv:2302.01721}, 2023.

\bibitem{rombach2022high}
Robin Rombach, Andreas Blattmann, Dominik Lorenz, Patrick Esser, and Bj{\"o}rn
  Ommer.
\newblock High-resolution image synthesis with latent diffusion models.
\newblock In {\em Proceedings of the IEEE/CVF Conference on Computer Vision and
  Pattern Recognition}, pages 10684--10695, 2022.

\bibitem{saharia2022photorealistic}
Chitwan Saharia, William Chan, Saurabh Saxena, Lala Li, Jay Whang, Emily
  Denton, Seyed Kamyar~Seyed Ghasemipour, Burcu~Karagol Ayan, S~Sara Mahdavi,
  Rapha~Gontijo Lopes, et~al.
\newblock Photorealistic text-to-image diffusion models with deep language
  understanding.
\newblock {\em arXiv preprint arXiv:2205.11487}, 2022.

\bibitem{saito2019pifu}
Shunsuke Saito, Zeng Huang, Ryota Natsume, Shigeo Morishima, Angjoo Kanazawa,
  and Hao Li.
\newblock Pifu: Pixel-aligned implicit function for high-resolution clothed
  human digitization.
\newblock In {\em Proceedings of the IEEE/CVF international conference on
  computer vision}, pages 2304--2314, 2019.

\bibitem{sanghi2022clip}
Aditya Sanghi, Hang Chu, Joseph~G Lambourne, Ye Wang, Chin-Yi Cheng, Marco
  Fumero, and Kamal~Rahimi Malekshan.
\newblock Clip-forge: Towards zero-shot text-to-shape generation.
\newblock In {\em Proceedings of the IEEE/CVF Conference on Computer Vision and
  Pattern Recognition}, pages 18603--18613, 2022.

\bibitem{seo2023let}
Junyoung Seo, Wooseok Jang, Min-Seop Kwak, Jaehoon Ko, Hyeonsu Kim, Junho Kim,
  Jin-Hwa Kim, Jiyoung Lee, and Seungryong Kim.
\newblock Let 2d diffusion model know 3d-consistency for robust text-to-3d
  generation.
\newblock {\em arXiv preprint arXiv:2303.07937}, 2023.

\bibitem{sun2021loftr}
Jiaming Sun, Zehong Shen, Yuang Wang, Hujun Bao, and Xiaowei Zhou.
\newblock Loftr: Detector-free local feature matching with transformers.
\newblock In {\em Proceedings of the IEEE/CVF conference on computer vision and
  pattern recognition}, pages 8922--8931, 2021.

\bibitem{trevithick2021grf}
Alex Trevithick and Bo Yang.
\newblock Grf: Learning a general radiance field for 3d representation and
  rendering.
\newblock In {\em Proceedings of the IEEE/CVF International Conference on
  Computer Vision}, pages 15182--15192, 2021.

\bibitem{varma2022attention}
Mukund Varma, Peihao Wang, Xuxi Chen, Tianlong Chen, Subhashini Venugopalan,
  and Zhangyang Wang.
\newblock Is attention all that nerf needs?
\newblock In {\em The Eleventh International Conference on Learning
  Representations}, 2022.

\bibitem{wang2022score}
Haochen Wang, Xiaodan Du, Jiahao Li, Raymond~A Yeh, and Greg Shakhnarovich.
\newblock Score jacobian chaining: Lifting pretrained 2d diffusion models for
  3d generation.
\newblock {\em arXiv preprint arXiv:2212.00774}, 2022.

\bibitem{wang2018pixel2mesh}
Nanyang Wang, Yinda Zhang, Zhuwen Li, Yanwei Fu, Wei Liu, and Yu-Gang Jiang.
\newblock Pixel2mesh: Generating 3d mesh models from single rgb images.
\newblock In {\em Proceedings of the European conference on computer vision
  (ECCV)}, pages 52--67, 2018.

\bibitem{wang2021neus}
Peng Wang, Lingjie Liu, Yuan Liu, Christian Theobalt, Taku Komura, and Wenping
  Wang.
\newblock Neus: Learning neural implicit surfaces by volume rendering for
  multi-view reconstruction.
\newblock {\em arXiv preprint arXiv:2106.10689}, 2021.

\bibitem{wang2021ibrnet}
Qianqian Wang, Zhicheng Wang, Kyle Genova, Pratul~P Srinivasan, Howard Zhou,
  Jonathan~T Barron, Ricardo Martin-Brualla, Noah Snavely, and Thomas
  Funkhouser.
\newblock Ibrnet: Learning multi-view image-based rendering.
\newblock In {\em Proceedings of the IEEE/CVF Conference on Computer Vision and
  Pattern Recognition}, pages 4690--4699, 2021.

\bibitem{wei2023taps3d}
Jiacheng Wei, Hao Wang, Jiashi Feng, Guosheng Lin, and Kim-Hui Yap.
\newblock Taps3d: Text-guided 3d textured shape generation from pseudo
  supervision, 2023.

\bibitem{wen2019pixel2mesh++}
Chao Wen, Yinda Zhang, Zhuwen Li, and Yanwei Fu.
\newblock Pixel2mesh++: Multi-view 3d mesh generation via deformation.
\newblock In {\em Proceedings of the IEEE/CVF international conference on
  computer vision}, pages 1042--1051, 2019.

\bibitem{wu2023multiview}
Chao-Yuan Wu, Justin Johnson, Jitendra Malik, Christoph Feichtenhofer, and
  Georgia Gkioxari.
\newblock Multiview compressive coding for 3d reconstruction.
\newblock {\em arXiv preprint arXiv:2301.08247}, 2023.

\bibitem{wu2017marrnet}
Jiajun Wu, Yifan Wang, Tianfan Xue, Xingyuan Sun, Bill Freeman, and Josh
  Tenenbaum.
\newblock Marrnet: 3d shape reconstruction via 2.5 d sketches.
\newblock {\em Advances in neural information processing systems}, 30, 2017.

\bibitem{xie2019pix2vox}
Haozhe Xie, Hongxun Yao, Xiaoshuai Sun, Shangchen Zhou, and Shengping Zhang.
\newblock Pix2vox: Context-aware 3d reconstruction from single and multi-view
  images.
\newblock In {\em Proceedings of the IEEE/CVF international conference on
  computer vision}, pages 2690--2698, 2019.

\bibitem{xie2020pix2vox++}
Haozhe Xie, Hongxun Yao, Shengping Zhang, Shangchen Zhou, and Wenxiu Sun.
\newblock Pix2vox++: Multi-scale context-aware 3d object reconstruction from
  single and multiple images.
\newblock {\em International Journal of Computer Vision}, 128(12):2919--2935,
  2020.

\bibitem{xu2022neurallift}
Dejia Xu, Yifan Jiang, Peihao Wang, Zhiwen Fan, Yi Wang, and Zhangyang Wang.
\newblock Neurallift-360: Lifting an in-the-wild 2d photo to a 3d object with
  360 $\{$$\backslash$deg$\}$ views.
\newblock {\em arXiv preprint arXiv:2211.16431}, 2022.

\bibitem{xu2022dream3d}
Jiale Xu, Xintao Wang, Weihao Cheng, Yan-Pei Cao, Ying Shan, Xiaohu Qie, and
  Shenghua Gao.
\newblock Dream3d: Zero-shot text-to-3d synthesis using 3d shape prior and
  text-to-image diffusion models.
\newblock {\em arXiv preprint arXiv:2212.14704}, 2022.

\bibitem{xu2019disn}
Qiangeng Xu, Weiyue Wang, Duygu Ceylan, Radomir Mech, and Ulrich Neumann.
\newblock Disn: Deep implicit surface network for high-quality single-view 3d
  reconstruction.
\newblock {\em Advances in neural information processing systems}, 32, 2019.

\bibitem{yagubbayli2021legoformer}
Farid Yagubbayli, Yida Wang, Alessio Tonioni, and Federico Tombari.
\newblock Legoformer: Transformers for block-by-block multi-view 3d
  reconstruction.
\newblock {\em arXiv preprint arXiv:2106.12102}, 2021.

\bibitem{yang2021robotic}
Daniel Yang, Tarik Tosun, Benjamin Eisner, Volkan Isler, and Daniel Lee.
\newblock Robotic grasping through combined image-based grasp proposal and 3d
  reconstruction.
\newblock In {\em 2021 IEEE International Conference on Robotics and Automation
  (ICRA)}, pages 6350--6356. IEEE, 2021.

\bibitem{yang2023contranerf}
Hao Yang, Lanqing Hong, Aoxue Li, Tianyang Hu, Zhenguo Li, Gim~Hee Lee, and
  Liwei Wang.
\newblock Contranerf: Generalizable neural radiance fields for
  synthetic-to-real novel view synthesis via contrastive learning.
\newblock {\em arXiv preprint arXiv:2303.11052}, 2023.

\bibitem{yang2018foldingnet}
Yaoqing Yang, Chen Feng, Yiru Shen, and Dong Tian.
\newblock Foldingnet: Point cloud auto-encoder via deep grid deformation.
\newblock In {\em Proceedings of the IEEE conference on computer vision and
  pattern recognition}, pages 206--215, 2018.

\bibitem{yu2021pixelnerf}
Alex Yu, Vickie Ye, Matthew Tancik, and Angjoo Kanazawa.
\newblock pixelnerf: Neural radiance fields from one or few images.
\newblock In {\em Proceedings of the IEEE/CVF Conference on Computer Vision and
  Pattern Recognition}, pages 4578--4587, 2021.

\bibitem{zeng2022lion}
Xiaohui Zeng, Arash Vahdat, Francis Williams, Zan Gojcic, Or Litany, Sanja
  Fidler, and Karsten Kreis.
\newblock Lion: Latent point diffusion models for 3d shape generation.
\newblock {\em arXiv preprint arXiv:2210.06978}, 2022.

\bibitem{zhang2023adding}
Lvmin Zhang and Maneesh Agrawala.
\newblock Adding conditional control to text-to-image diffusion models, 2023.

\bibitem{zhang2022nerfusion}
Xiaoshuai Zhang, Sai Bi, Kalyan Sunkavalli, Hao Su, and Zexiang Xu.
\newblock Nerfusion: Fusing radiance fields for large-scale scene
  reconstruction.
\newblock In {\em Proceedings of the IEEE/CVF Conference on Computer Vision and
  Pattern Recognition}, pages 5449--5458, 2022.

\bibitem{zhou2023sparsefusion}
Zhizhuo Zhou and Shubham Tulsiani.
\newblock Sparsefusion: Distilling view-conditioned diffusion for 3d
  reconstruction.
\newblock In {\em CVPR}, 2023.

\bibitem{zuffi2018lions}
Silvia Zuffi, Angjoo Kanazawa, and Michael~J Black.
\newblock Lions and tigers and bears: Capturing non-rigid, 3d, articulated
  shape from images.
\newblock In {\em Proceedings of the IEEE conference on Computer Vision and
  Pattern Recognition}, pages 3955--3963, 2018.

\bibitem{zuffi20173d}
Silvia Zuffi, Angjoo Kanazawa, David~W Jacobs, and Michael~J Black.
\newblock 3d menagerie: Modeling the 3d shape and pose of animals.
\newblock In {\em Proceedings of the IEEE conference on computer vision and
  pattern recognition}, pages 6365--6373, 2017.

\end{thebibliography}
}

\end{document}